\documentclass{article}

\usepackage{listings}
\usepackage[frozencache=true, cachedir=minted-cache]{minted}
\usepackage{tcolorbox}

\usepackage{microtype}
\usepackage{graphicx}
\usepackage{booktabs} 

\usepackage{hyperref}

\usepackage{amsmath}
\usepackage{amssymb}
\usepackage{mathtools}
\usepackage{amsthm}

\usepackage[capitalize,noabbrev]{cleveref}

\theoremstyle{plain}

\theoremstyle{definition}

\theoremstyle{remark}

\usepackage[textsize=tiny]{todonotes}

\usepackage[normalem]{ulem}
\usepackage{subcaption}
\usepackage{graphicx}
\usepackage{csvsimple-l3}
\usepackage{pgfplotstable}
\usepackage{pdfpages}
\usepackage{svg}
\usepackage{enumitem}

\usepackage[accepted]{icml2024}

\icmltitlerunning{In-Context Reinforcement Learning for Variable Action Spaces}

\begin{document}

\twocolumn[
\icmltitle{In-Context Reinforcement Learning for Variable Action Spaces}



\icmlsetsymbol{attink}{*}

\begin{icmlauthorlist}
\icmlauthor{Viacheslav Sinii}{tink,inno}
\icmlauthor{Alexander Nikulin}{airi,mipt,attink}
\icmlauthor{Vladislav Kurenkov}{airi,inno,attink}
\icmlauthor{Ilya Zisman}{airi,skol,attink}
\icmlauthor{Sergey Kolesnikov}{tink}
\end{icmlauthorlist}

\icmlaffiliation{tink}{Tinkoff, Moscow, Russia}
\icmlaffiliation{airi}{AIRI, Moscow, Russia}
\icmlaffiliation{inno}{Innopolis University}
\icmlaffiliation{mipt}{MIPT}
\icmlaffiliation{skol}{Skoltech, Moscow, Russia}

\icmlcorrespondingauthor{Viacheslav Sinii}{v.siniy@tinkoff.ai}

\icmlkeywords{Machine Learning, Reinforcement Learning, In-Context Learning, Variable Action Spaces, Transformers}

\vskip 0.3in
]



\printAffiliationsAndNotice{\icmlEqualContribution} 

\begin{abstract}
Recently, it has been shown that transformers pre-trained on diverse datasets with multi-episode contexts can generalize to new reinforcement learning tasks in-context. A key limitation of previously proposed models is their reliance on a predefined action space size and structure. The introduction of a new action space often requires data re-collection and model re-training, which can be costly for some applications. In our work, we show that it is possible to mitigate this issue by proposing the Headless-AD model that, despite being trained only once, is capable of generalizing to discrete action spaces of variable size, semantic content and order. By experimenting with Bernoulli and contextual bandits, as well as a gridworld environment, we show that Headless-AD exhibits significant capability to generalize to action spaces it has never encountered, even outperforming specialized models trained for a specific set of actions on several environment configurations. Implementation is available at: \url{https://github.com/corl-team/headless-ad}.
\end{abstract}

\section{Introduction}
\label{intro}

\begin{figure}[h]
    \vskip 0.2in
    \centering
    \includegraphics[width=\columnwidth]{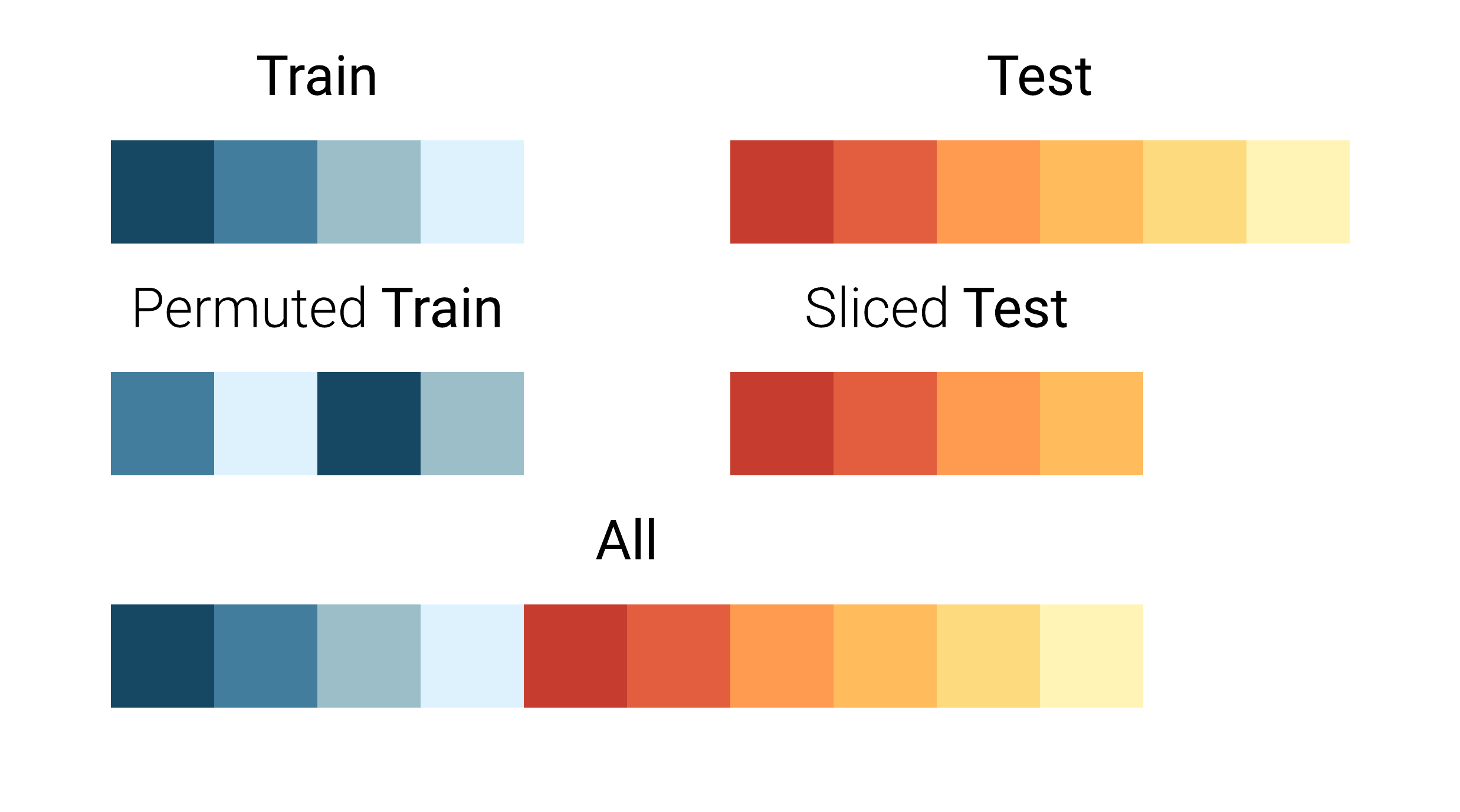}
    \caption{\textbf{Variable Action Spaces:} We consider four types of novel action spaces different from the one used during training. \emph{Permuted Train Actions} maintains the action set contents but reorders its elements. \emph{Test Actions} introduces a completely new action set with an increased size. It is important to consider that some models may be architecturally limited to a fixed action set size. To evaluate the performance of such models on unseen actions, we adjust the size of a new set to be compatible with the model output. Therefore, we slice the first actions from the \emph{Test Actions} set. Lastly, a new action space might include both the seen \emph{Train} and unseen \emph{Test} actions, depicted as the \emph{All Actions} set.}
    \label{fig:action_sets_viz}
    \vskip -0.2in
\end{figure}

\begin{figure*}[t]
    \vskip 0.2in
    \centering
    \includegraphics[width=\textwidth]{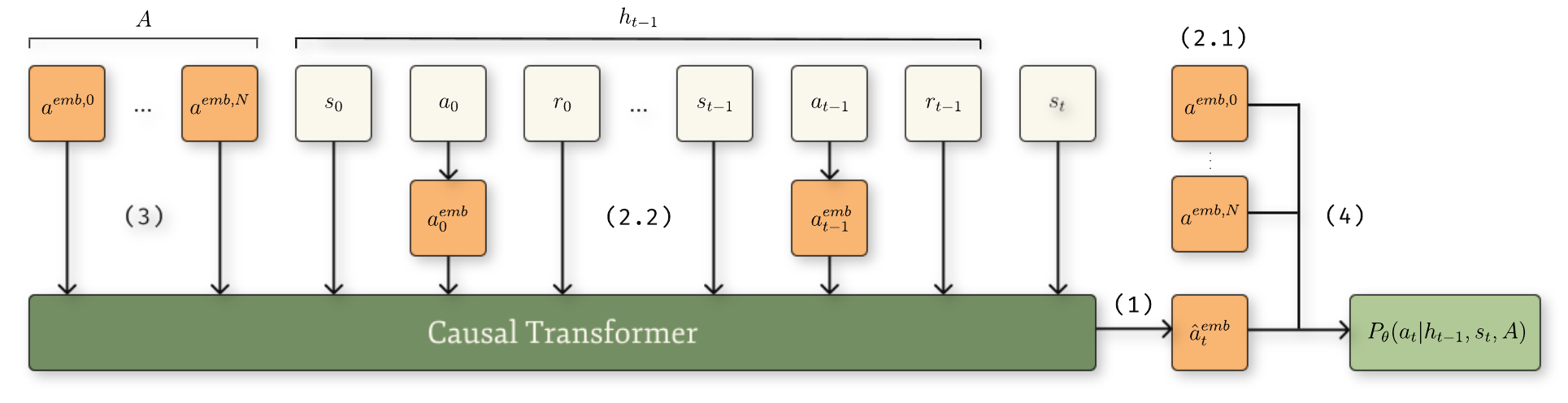}
    \caption{\textbf{Headless-AD Architecture:} Compared to AD, Headless-AD introduces four new components. (1) We remove the output linear head, making the model directly predict the action embedding. That allows us to avoid a direct connection between the model and action space size, contents and ordering. (2.1) At each training step, we generate random action embeddings for each action in the action set. (2.2) We convert actions in the context into their embeddings and pass them as the model input. This prepares the model for unseen actions, forcing it to infer action semantics from the context. (3) As the model loses prior knowledge about action space structure, we pass the generated action embeddings as a prompt to aid the model in sensible action selection. (4) We convert a prediction vector into a distribution over actions based on the similarities between the prediction and previously generated action embeddings. To increase the probability of correct actions, we use contrastive loss instead of cross-entropy.}
    \label{fig:model_vis}
    \vskip -0.2in
\end{figure*}

The transformer architecture, first introduced by \citet{vaswani2017attention}, has been widely adopted in key areas of machine learning, including natural language processing \citep{radford2018improving, devlin2018bert}, computer vision \citep{dosovitskiy2020image} and sequential decision-making \citep{chen2021decision}. One major feature of transformers is in-context learning (ICL), which makes it possible for them to adapt to new tasks after extensive pre-training \citep{brown2020language, liu2023pre}. Recent developments, such as Algorithm Distillation (AD) by \citet{laskin2022context} and Decision Pretrained Transformer (DPT) by \citet{lee2023supervised}, have successfully employed transformer ICL abilities in sequential decision-making. These models are capable of predicting the next action based on a query state and history of environment interactions, which inform them about task objectives and environment dynamics. While effective at generalizing across various reward distributions \citep{laskin2022context, lee2023supervised} and transition functions \citep{raparthy2023generalization}, their adaptability to new action spaces remains unexplored and limited by architectural constraints.

Creating models that can adapt to new action spaces is essential for building the foundation of decision-making systems in order to enable large-scale pretraining across various environments and address real-world problems \citep{jain2020generalization, chandak2020lifelong, london2020offline, jain2021know}. With this in mind, our research focuses on variable discrete action spaces, with the notion of variability illustrated in the \Cref{fig:action_sets_viz}. In our study, we reveal the limitations of the Algorithm Distillation model from prior work, such as its diminished performance upon changes in action semantics as well as architectural constraints when handling varying action space sizes (see \Cref{fig:bad_ad}). 

Our solution, \textbf{Headless-AD}, is an architecture and training methodology tailored to effective generalization on new action spaces. We employ an approach similar to Wolterpinger \citep{dulac2015deep} and Headless-LLM \citep{godey2023headless} by encoding actions with random embeddings \citep{kirsch2023towards} and directly predicting these embeddings. This way, we remove the direct connection between the model output layer and the action space structure. 

Through experiments using Bernoulli and contextual bandits, and a darkroom environment with changing action spaces, we demonstrate that Headless-AD is capable of matching the performance of the original data generation algorithm and scaling to action spaces up to $\textbf{5x}$ larger than those seen during training. We also observed that Headless-AD can even outperform AD when they are both trained for the same action space, especially when evaluated on larger action sets. To summarize, our contributions are as follows:
\begin{itemize}
    \item We show that AD struggles with generalization on novel action spaces (\Cref{section:ad_struggles}).
    \item We extend AD with a modified model architecture and a training strategy, called \textbf{Headless-AD}, for it to acquire the ability to adapt to new discrete action spaces (\Cref{section:headless_ad}). We demonstrate the strong generalization capabilities of Headless-AD on Bernoulli and contextual bandits, and darkroom environments (\Cref{section:experiments}).
    \item We perform ablations on the loss and the prompt format to highlight the importance of Headless-AD's design choices (\Cref{section:ablations}).
\end{itemize}
\vphantom{loh}\\

\section{Algorithm Distillation Struggles with Novel Action Spaces}
\label{section:ad_struggles}

Algorithm Distillation (AD) \citep{laskin2022context} is a transformer model trained to autoregressively predict the next action given the history of previous environment interactions and a current state. Formally, the history is defined as:
\[
h_t = \left(o_0, a_0, r_0, \dots, o_t, a_t, r_t\right),
\]
where $o$ are the observations, $a$ are the actions and $r$ are the rewards. The probabilities of each action are given by a model $P_\theta(A=a^{(n)}_t|h_{t-1}, o_t)$, where $n$ is the action index, $t$ is a timestamp and $\theta$ are the model weights. 
AD is pretrained on data logged by a training agent, and the context size should be sufficiently large to span multiple episodes in order to capture policy improvement. This way, AD learns an improvement operator that increases performance entirely in-context when applied to novel tasks.

\begin{figure}[!t]
    \vskip 0.2in
    \centering
    \includegraphics[width=\columnwidth]{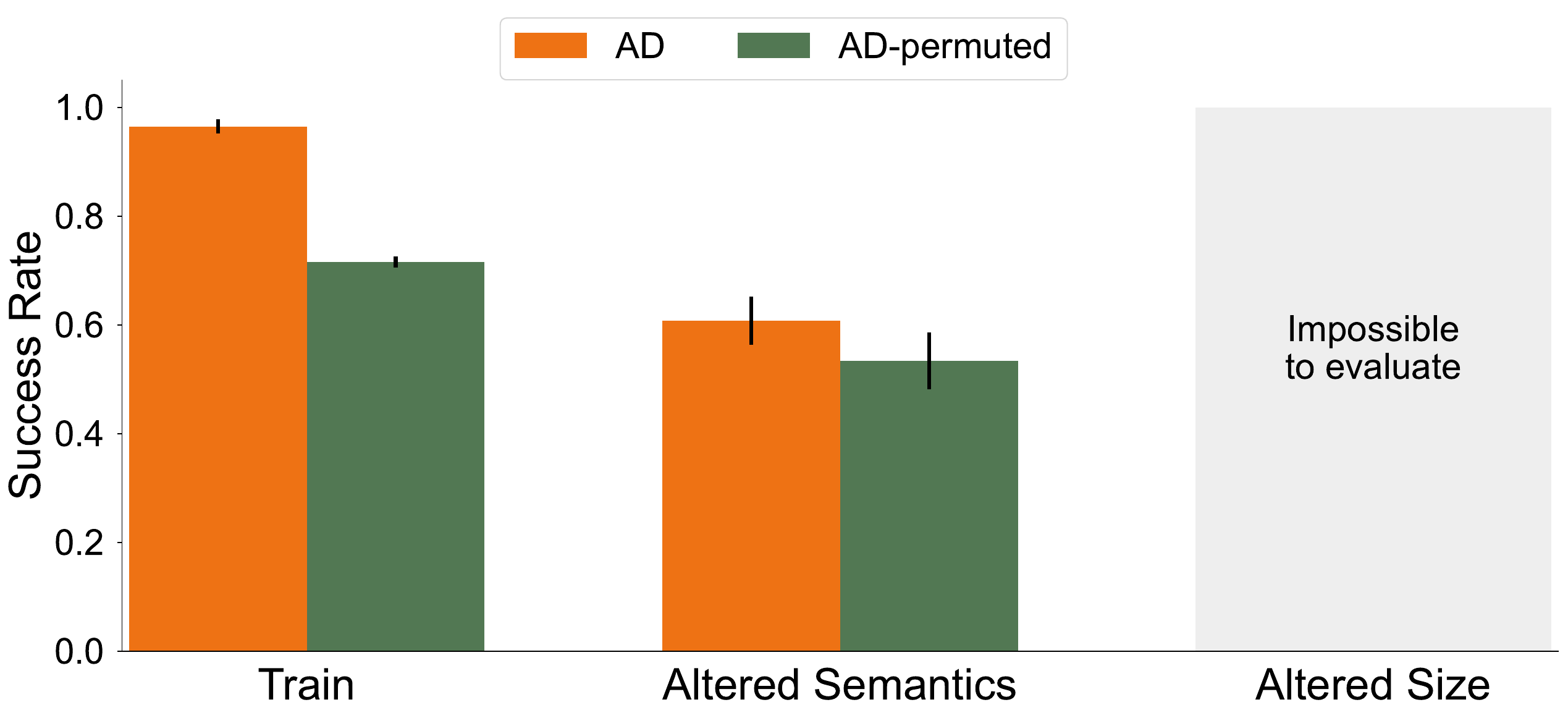}
    \caption{\textbf{Algorithm Distillation Struggles with Novel Action Spaces:} Despite its good results on the train action set, AD's performance diminishes when the action semantics change, either due to a permutation or substitution. It is important to note that augmenting the training data with permuted action sets does not lead to increased performance, signifying that action set invariance should be enforced from a model design standpoint. Additionally, it is impossible to apply a trained AD model to a larger action set. On the graph, the bars are the success rate values on the Darkroom environment (described in \Cref{sec:exp_gridworld}) obtained after evaluating each of the action sets visualized in \Cref{fig:action_sets_viz}, averaged over 5 runs. \textit{Altered Semantics} aggregate the values from the \textit{Permuted Train Actions} and \textit{Sliced Test Actions} sets. \textit{Altered Size} aggregates the values from \textit{Test Actions} and \textit{All Actions}. See \cref{sec:exp_gridworld} for more information about the construction of the action sets.}
    \label{fig:bad_ad}
    \vskip -0.2in
\end{figure}

The model output is a probability distribution across the action set, derived from a linear projection and a softmax function. As highlighted in \Cref{fig:bad_ad}, this structure causes a fundamental limitation in AD's adaptability to new action spaces, as the output dimension is predetermined. To accommodate an action set of a different size than the one used during training, the model's final layer must be redefined and the model retrained. Moreover, even with a constant action space size, the model's efficacy diminishes if the action semantics are altered. The reason for this is the classifier nature of the model, which associates each dimension with a particular meaning of an action. Since augmenting the dataset with permuted action sets does not lead to improvement, it signifies that action set invariance should be enforced from a model design standpoint.
\section{Headless-AD}
\label{section:headless_ad}

\begin{figure}[t]
    \vskip 0.2in
    \centering
    \includegraphics[width=\columnwidth]{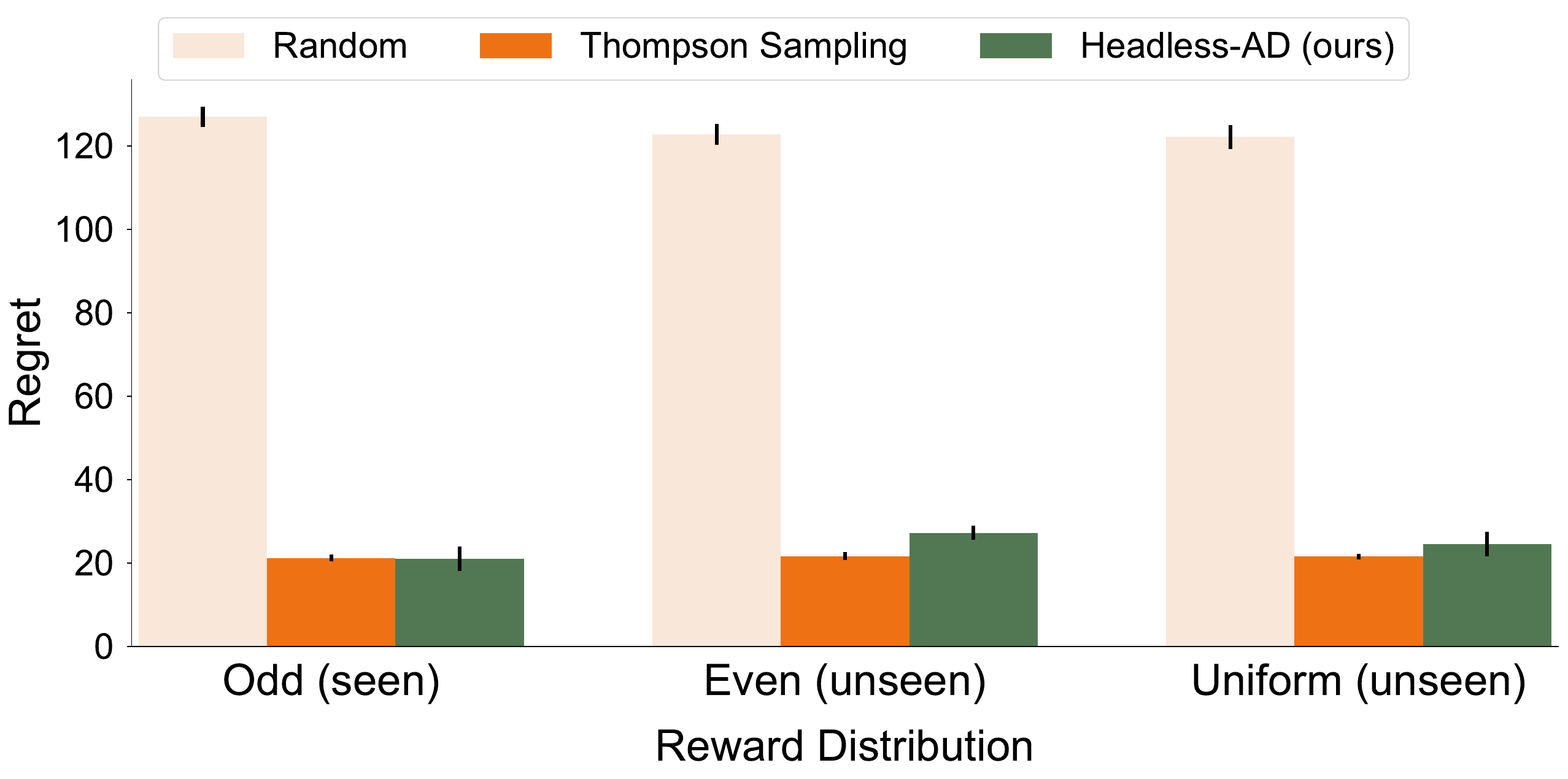}
    \caption{\textbf{Algorithm Regret under Variable Reward Distributions in Bernoulli Bandit:} The graph compares regret for Random, Thompson Sampling, and Headless-AD across distinct reward distributions in the Bernoulli Bandit environment, averaged from five seeds. During training, the high reward was $95\%$ more likely to distribute across the odd arms. During testing, it either switched to the even arms or a uniform distribution. Note that Headless-AD maintains high performance in all configurations, proving its ICL capabilities at generalizing to novel tasks, represented by changes in reward distribution. Data is aggregated from bandit problems with $4-20$ arms, reflecting the training conditions.}
    \label{fig:bandit_new_regrets}
    \vskip -0.2in
\end{figure}

To mitigate the action space limitations of AD, we propose Headless-AD, a new architecture that improves on AD by omitting the final linear layer and incorporating three key modifications. The Headless-AD architecture and data flow are visualized in \Cref{fig:model_vis}.


\textbf{Random Action Embeddings:} To remove the dependence of the model on the pretrained action embeddings, we employed a dynamic mapping function $g:~\mathbb{A}~\rightarrow~\mathbb{R}^n$, which produces a unique random encoding for each action in a batch at the start of every training step. The mapping is shared across all batch instances and is consistent along the context sequence, i.e., actions with index $i$ in their respective action sets will all be mapped to the same embedding. During inference, we generated a single set of embeddings at the beginning and used them throughout the evaluation.

The core intuition behind employing random action embeddings is to eliminate any prior knowledge about the structure of the action space within our model. This approach stems from our observation that using learnable embeddings for actions becomes impractical when encountering new actions not seen during training. A new action would lack a pre-trained embedding, and assigning an arbitrary embedding could introduce an undesirable domain shift, as the model would not recognize it. Moreover, allowing the model to learn new embeddings on-the-fly would necessitate extra gradient steps, diverging from our goal of maintaining a zero-shot learning framework. The usage of random action embeddings ensures that the model does not depend on extracting any information from the embeddings themselves but rather relies on interpreting the context provided by historical interactions with the environment. Moreover, employing random embeddings enhances data variety, which has been demonstrated to boost in-context learning for RL agents \citep{kirsch2023towards, lu2023structured}.

\begin{figure*}[t]
    \vskip 0.2in
    \centering
    \includegraphics[width=0.95\textwidth]{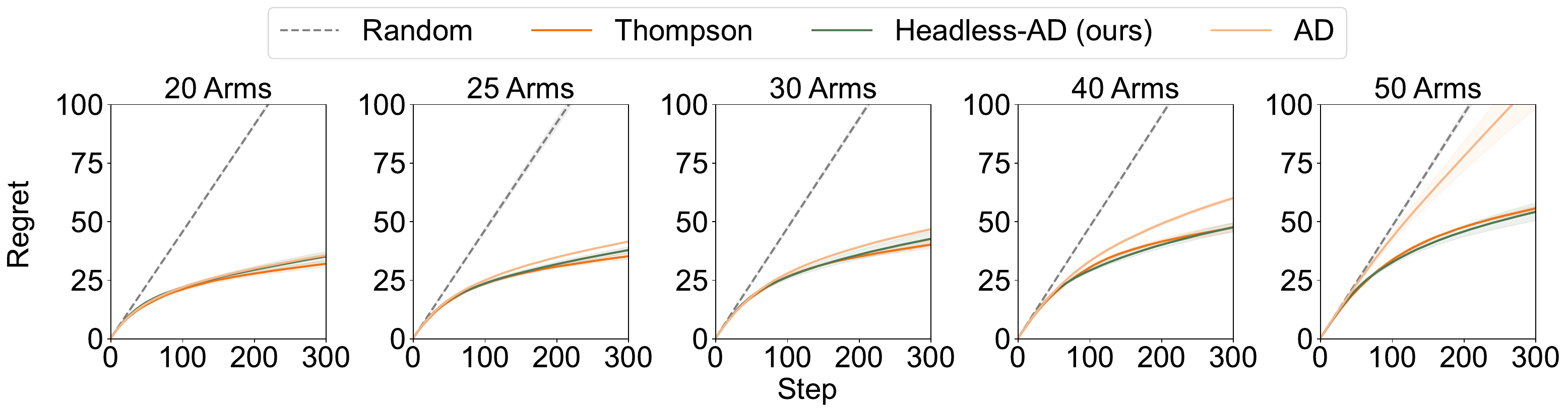}
    \caption{\textbf{Algorithm Regret under Increasing Amount of Arms in Bernoulli Bandit:} This series of plots shows the regret of Thompson Sampling, AD, and Headless-AD algorithms over evaluation steps in environments with $20-50$ arms, averaged from five seeds with $100$ bandits each. Although Headless-AD has been trained on bandits with up to $20$ arms, it performs well, matching or outperforming other algorithms in larger arm settings without additional training. Note that AD was retrained from scratch for each task with a different number of arms.}
    \label{fig:bandit_together_curves}
    \vskip -0.2in
\end{figure*}

We further refined this strategy by constraining the random embeddings to lie on a unit sphere and ensuring their orthogonality (see \Cref{appendix:orthonormal_vectors}). The choice of a unit sphere normalizes the scale of the embeddings, while orthogonality allows the model to independently adjust the probability assigned to each action. This condition is crucial for preventing unintended probability mass allocation to multiple actions when the model's prediction vector aligns with one embedding vector. A similar concept is explored by \citep{elhage2022superposition} in the context of feature interference.

\textbf{Direct Prediction of Action Embeddings:} The model output is modified to yield an action embedding $\hat{a}^{emb}_t$ rather than a probability distribution over actions. This alteration makes the model independent of action set size and order, granting it permutation invariance. The InfoNCE Contrastive Loss \citep{oord2018representation}, diverging from its usual role in representation learning \citep{jaiswal2020survey}, serves as a regression objective to reinforce the similarity between the model prediction and the subsequent action in the data. All other actions are treated as negative samples. Thus, the objective is 
\[
L = -\mathop{\mathbb{E}}\left[{\log{\frac{e^{f(\hat{a}^{emb}_t, a^{emb}_t) / \tau}}{\sum_{a \in A}{e^{f(\hat{a}^{emb}_t, a^{emb}) / \tau}}}}}\right],
\]
where $a^{emb} = g(a)$ and $\tau$ is a temperature parameter. We used dot-product as the similarity function $f$. 

\textbf{Action Set Prompt:} To address the model's lack of awareness of the action space structure caused by the two previous changes, we prepend the input with a sequence of embeddings for all available actions.

The modified input format is thus represented as
\[
h_t = \left(a^{emb,0}, \dots, a^{emb,N}, o_{<t}, a^{emb}_{<t}, r_{<t}, o_t\right),
\]
where $N$ is the action set size. An illustrative code snippet with Headless-AD's training procedure can be found in \Cref{appendix:pseudo_code}.
\newline
\newline
We suggest two methods to select actions during inference: (1) nearest neighbor selection:
\[
a = \arg \max_{a \in A} f(\hat{a}^{emb}, a^{emb})
\]
and (2) probabilistic sampling based on the similarity to the predicted action embedding:
\[
P(a_i) = \frac{e^{f(\hat{a}^{emb}, a^{emb}_i)}}{\sum_{a \in A}{e^{f(\hat{a}^{emb}, a^{emb}}})}.
\]
We treat the specific choice of method as a hyperparameter.
\section{Experiments}
\label{section:experiments}

As Headless-AD extends and improves on AD, we checked it in two different aspects. Firstly, it should maintain In-Context Learning abilities and thus generalize well to new tasks. Secondly, it should show high performance on action spaces different from the one seen during training. All of the following environments are designed specifically to check both of the above aspects. In our experiments, we used the TinyLLaMa \citep{zhang2024tinyllama} implementation of the transformer model and AdamW optimizer \citep{loshchilov2017decoupled}. All environment specific hyperparameters are listed in \Cref{appendix:hypers}.

\subsection{Bernoulli Bandit}
\label{eval:bernoulli_bandit}

\begin{figure*}[t]
    \vskip 0.2in
    \centering
    \includegraphics[width=\textwidth]{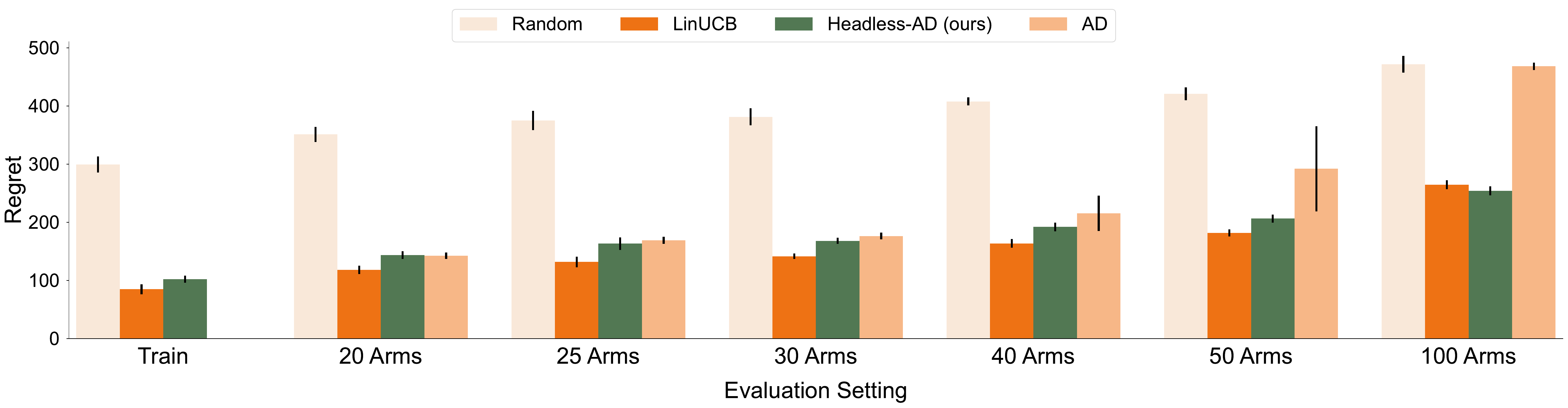}
    \caption{\textbf{Contextual Bandit Regret Comparison:} The \textit{Train} set consists of trajectories of LinUCB trained for $300$ steps on contextual bandits with $4-20$ arms. All models are also evaluated for $300$ steps. The results are averaged over $5$ seeds, each seed containing $100$ environments. AD requires retraining on each new action set and, while showing good performance on lower space sizes, it fails to converge on larger ones. Due to its variable-size action sets, AD was not tested on the \textit{Train} set. Conversely, Headless-AD, trained exclusively on the \textit{Train} set, is successful at both learning effectively within this environment and generalizing to new action sets. }
    \label{fig:contextual}
    \vskip -0.2in
\end{figure*}

\textbf{Motivation:}
This experiment checked Headless-AD's abilities on a toy task, where the environment did not have a notion of state and returned binary rewards.

\textbf{Setup 1:} In our first experiment, we examined the model's robustness to distributional shifts in rewards. We used a Bernoulli bandit where each arm is associated with a mean $\mu$ and the reward after pulling an arm $i$ is generated by $\text{Bernoulli}(\mu)$. The training dataset consisted of bandits with $4-20$ arms, i.e., different action set sizes. Additionally, to evaluate the ICL capabilities of models, the reward distribution over the arms differed between train and test \citep{laskin2022context}. Specifically, the training data consisted of bandits where in $95\%$ of cases the odd-numbered arms were assigned a random $\mu \in [0.5, 1]$ and even-numbered arms received $\mu \in [0, 0.5]$. For other $5\%$ of bandits, the ranges were swapped between odd and even arms. The test distribution also consisted of bandits with $4-20$ arms, but the reward distribution was switched either to even arms or was uniform, i.e., all arms were assigned $\mu \in [0, 1]$. Learning histories were generated using Thompson Sampling algorithm for a total of $10,000$ bandit instances with $300$ steps each. The evaluation was performed on $100$ bandits in each reward distribution, included $5$ seeds, and the algorithms were rolled out for $300$ steps.

\textbf{Results and Discussion 1:} As depicted in \Cref{fig:bandit_new_regrets}, the Headless-AD model demonstrates strong generalization capabilities by almost reaching the performance results set by the traditional Thompson Sampling algorithm under each test distribution. 

\textbf{Setup 2:} In our second experiment, we evaluated the transferability of Headless-AD to new action set sizes. Training distribution remained the same as in the previous experiment. Each evaluation dataset consisted of a fixed amount of $20$, $25$, $30$, $40$ and $50$ arms and a uniform distribution of rewards. AD was trained on fixed-size bandits corresponding to each evaluation dataset. The training and evaluation reward distributions were the same as for Headless-AD.

\textbf{Results and Discussion 2:} \Cref{fig:bandit_together_curves} shows that Headless-AD can effectively maintain its performance as the action space grows, without necessitating any retraining. Moreover, Headless-AD even outperforms a specially trained AD, especially on larger action sets. Note that Headless-AD's performance curves resemble TS's performance curves, signifying that it has indeed learned a policy improvement operator generic enough to apply to unseen action set sizes.

\subsection{Contextual Bandit}
\label{eval:contextual_bandit}

\textbf{Motivation:} To validate the sustained performance of Headless-AD, we progressed to a more complex Multi-Armed Bandit (MAB) extension that integrated states and real-valued rewards.

\textbf{Setup:} Each time step presented a context with arm features modeled as two-dimensional vectors, and rewards were generated with a standard deviation $\sigma = 1$. The data were created using a LinUCB \cite{li2010contextual} algorithm trained over $300$ steps, on bandits with $4-20$ arms. We used $100$ contextual bandits, $5$ seeds and $300$ evaluation steps.

\textbf{Results and Discussion:} As shown in \Cref{fig:contextual}, Headless-AD's performance is on par with LinUCB across varied arm counts. While AD also reaches the performance of LinUCB on lower space sizes, it has problems converging on larger action space sizes, in addition to requiring a specific retraining. This serves to highlight the advantages of Headless-AD for use in even more complex environments than toy Bernoulli bandits.

\begin{figure*}[t]
    \vskip 0.2in
    \centering
    \includegraphics[width=0.95\textwidth]{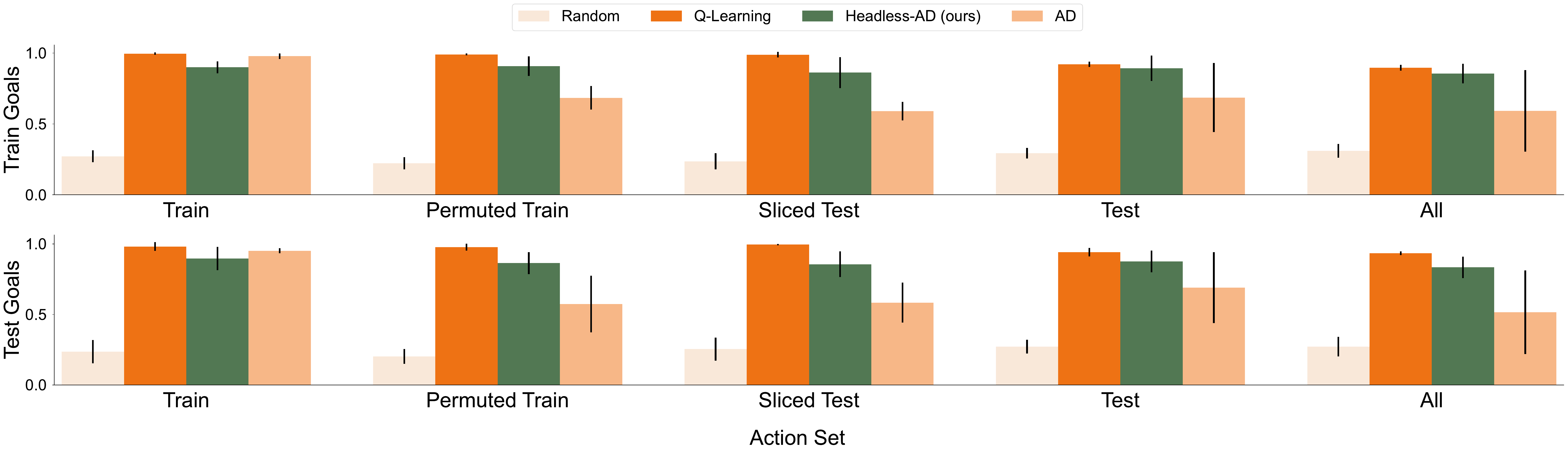}
    \caption{\textbf{Darkroom Environment Success Rate:} The chart displays mean success rates and their standard deviations from five training seeds, comparing performance in fixed and variable action spaces, along with the models' adaptability to new goals. \textit{Train Actions} refers to the fixed-size action set used for training. \textit{Test Actions} include exclusively unseen actions, while \textit{All Actions} combine both \textit{Train} and \textit{Test}, with set sizes expanded to $75$ and $125$ respectively. Since the output dimension changes, AD requires retraining. To assess AD's adaptability to the changed action semantics, we either permute \textit{Train Actions} or replace it with the first $50$ actions from the \textit{Test} set. In this case, AD does not require retraining, but it still exhibits diminished performance. Conversely, Headless-AD, while trained solely on \textit{Train Actions}, delivers strong and stable performance across all action set variants, and even surpasses specially trained AD on larger set sizes.}
    \label{fig:gridworld_bars}
    \vskip -0.2in
\end{figure*}

\subsection{Darkroom}
\label{sec:exp_gridworld}

\textbf{Motivation:} In this experiment, we delve into a more sophisticated Markov Decision Process (MDP) framework, constructing five distinct action spaces aimed at demonstrating the architectural constraints of AD. We then show how Headless-AD's architecture is engineered to navigate the complexities presented by each of these diverse action sets.

\textbf{Setup:} The Darkroom environment, inspired by \citet{chevalier2018minimalistic} and \citet{jain2020generalization}, consists of a $N \times N$ grid where the agent needs to reach a specific cell for a reward. In our experiment, the action space consisted of $3$-step sequences of $5$ atomic actions: \textit{up}, \textit{down}, \textit{left}, \textit{right}, \textit{noop}. As a result, the environment offered $5^3$ possible actions. The agent earned a reward of $1$ if the trajectory induced by the action sequence passed through a goal cell, after which the episode finished. Otherwise, the reward was $0$. As an observation, the environment offered only the current coordinates of the agent, so the goal information could only be obtained from the agent's memory. We divided the goals into disjoint sets used for training and testing in order to evaluate Headless-AD's in-context learning abilities. Furthermore, the action set was randomly split into train and test sets, each including $50$ and $75$ actions respectively, to create five distinct spaces for assessing various generalization aspects. These action sets are visualized in \Cref{fig:action_sets_viz}.

\textbf{Train Actions.} Comprising the training split, this set represented the actions encountered during model training.

\textbf{Test Actions.} Comprising the test split, this set assessed Headless-AD's generalization on novel and larger action sets.

\textbf{All Actions.} Combining both training and testing actions, this set contained $125$ actions and was $2.5$ times larger than the training set. Its aim was to challenge Headless-AD to effectively integrate seen and unseen actions.

\textbf{Permuted Train Actions.} Shuffled training set, meant to test the model's adaptability to reordered action spaces that comprised of the same actions.

\textbf{Sliced Test Actions.} Tailored to match the training set size, this set contained a slice of the first $50$ actions from the test set. Its aim was to check the models' generalization abilities on unseen actions while maintaining the action set size.

In scenarios where the action space exceeded the size of the training set, we analyzed the performance of an AD model retrained from scratch. The data generation algorithm was Q-learning, executed over $200$ episodes for each environment. Further details on model hyperparameters are available in \Cref{appendix:hypers}.

\textbf{Results and Discussion:} \Cref{fig:gridworld_bars} illustrates the generalization abilities of both AD and Headless-AD. First, note that both models maintain their performance when transitioning to novel tasks, thus fulfilling their purpose as ICL-RL models. However, this environment was mainly designed to challenge the models' generalization abilities on novel action spaces. Following the scope of the action space novelty we set earlier, we studied the models' ability to address changed action semantics and variable action set sizes. 

While AD achieves high performance on the train action set, its limitations become evident when the action space is changed. The first limitation is AD's action set size constraint. The linear layer at the end of its network fixes the size of the output dimension, something that cannot be modified once the model is trained. Increasing the amount of options, as was done in the \textit{Test Actions} and \textit{All Actions} sets, requires reinitializing the output dimension and thus retraining the model, which demands additional time and resources. In contrast, once trained, Headless-AD easily adapts to changes in the action set size without losing performance. 

\begin{table*}[!t]
\centering
\caption{\textbf{Ablations:} Table compares the performance of Headless-AD with its ablated versions. Columns 'Bandit' and 'Darkroom' show the performance averaged along all action sets. Bernoulli Bandit performance is normalized, where $0$ denotes a random agent and $1$ denotes the Thompson Sampling algorithm. Columns 'Bandit. Arms Used' and 'Darkroom. Arms Used' show the amount of actions tried by the model during evaluation, also averaged along all the action sets. Columns 'Bandit. \textit{N} Arms' show the performance of models on the Bernoulli Bandit environment for each respective number of arms during evaluation without averaging. The results are aggregated over $5$ random seeds. As the table shows, changing each component of Headless-AD's architecture greatly damages the model's ability either to utilize the action set effectively or to perform well.}
\label{table:ablations}
\resizebox{\textwidth}{!}{%

\begin{filecontents*}{tables/ablations.csv}
Bandit,Bandit. Arms Used,Darkroom,Darkroom. Acts Used,Bandit. 20 Arms,Bandit. 30 Arms,Bandit. 40 Arms,Bandit. 50 Arms
0.98 ± 0.02,19.27 ± 9.26,0.87 ± 0.03,57.21 ± 21.89,0.97 ± 0.02,0.98 ± 0.02,1.0 ± 0.02,1.02 ± 0.04
0.81 ± 0.12,22.28 ± 13.11,0.83 ± 0.03,54.69 ± 18.63,0.91 ± 0.03,0.81 ± 0.05,0.66 ± 0.03,0.55 ± 0.03
0.63 ± 0.08,2.4 ± 0.22,0.25 ± 0.04,70.0 ± 29.15,0.6 ± 0.22,0.64 ± 0.27,0.7 ± 0.22,0.8 ± 0.25
0.79 ± 0.14,23.78 ± 12.54,0.81 ± 0.03,55.72 ± 19.29,0.85 ± 0.01,0.74 ± 0.03,0.62 ± 0.04,0.52 ± 0.04
\end{filecontents*}
\pgfplotstabletypeset[
    font=\scriptsize,
    my multistyler/.style 2 args={
        @my multistyler/.style={columns/##1/.append style={#2}},
        @my multistyler/.list={#1}
    },
    create on use/row_names/.style={
        create col/set list={
            Headless-AD,
            Prompt Ablation,
            Loss Ablation,
            Embed Ablation
            },
    },
    columns/row_names/.style={string type},
    every head row/.style={before row=\toprule, after row=\midrule},
    every last row/.style={after row=\bottomrule},
    col sep=comma,
    columns={row_names,Bandit,Bandit. Arms Used,Darkroom,Darkroom. Acts Used,Bandit. 20 Arms,Bandit. 30 Arms,Bandit. 40 Arms,Bandit. 50 Arms},
    my multistyler={row_names, Bandit,Bandit. Arms Used,Darkroom,Darkroom. Acts Used,Bandit. 20 Arms,Bandit. 30 Arms,Bandit. 40 Arms,Bandit. 50 Arms}{string type},
    columns/row_names/.append style={column name=Setting, column type={l}},
]{tables/ablations.csv}
}
\vspace{-10pt}
\end{table*}

The second limitation is AD's reliance on a stationary action space structure. Due to the classifier nature of AD's network, it learns to associate each output dimension with a specific action meaning. When action semantics change either due to a permutation of seen actions, as in \textit{Permuted Train Actions}, or due to a substitution of completely new actions, as in \textit{Sliced Test Actions}, AD's performance degrades. We checked whether this problem may be solved by biasing the training distribution to have a structure resembling the one during testing. We trained AD on permuted action sets while leaving the data and the model architecture the same. However, as one can see in \Cref{fig:bad_ad}, this training procedure resulted only in a slightly decreased performance of AD in the \textit{Permuted Train} setting. An additional graph depicting the performance on each of the (action set, goal type) pairs can be found in \Cref{appendix:ad_perm}. Meanwhile, \textit{Permuted Train Actions} does not pose a challenge for Headless-AD as, by design, it is invariant to specific action order. Most importantly, Headless-AD maintains its performance even when completely new actions are introduced, despite having never seen them in training. We attribute this feature to our success in making the model infer action meaning from the context.

\section{Ablations}
\label{section:ablations}


The hyperparameters tuned for each ablation can be found in \Cref{appendix:hypers}.

\subsection{Action Set Prompt}
We assessed the impact of omitting the action embedding enumeration from the model context.


\Cref{table:ablations} reveals that removing the action set prompt did not significantly alter the number of unique actions attempted by each model. This suggests that models without the prompt continue to sample a diverse range of actions, comparable to the original setup. However, the presence of the prompt did affect the performance. We hypothesize that,
without the action set prompt, the model lacked explicit knowledge of the action space, which may have led to a more randomized sampling within the embedding space. Conversely, with the prompt in place, the model could directly target specific action embeddings.

The absence of the prompt was more detrimental in the Bernoulli Bandit environment, where each decision has a direct impact on the final outcome due to the environment's single-episode structure. However, in a multi-episode environment such as the Darkroom, the early lack of action space information becomes less impactful over time, as the model eventually encounters all actions through the context. This divergence highlights the prompt's critical role in enabling the model to make informed choices in environments where each option has immediate and lasting consequences. Note that the impact of action set prompt ablation on the performance got more pronounced as the number of arms grows.

\subsection{Contrastive Loss}
In principle, the prediction of action embedding can be facilitated by various ways. Here, we considered the performance of a model that was asked to directly copy the corresponding embedding from its context to the output. To achieve it, we used mean squared error (MSE) loss instead of contrastive loss.

In this case, probabilistic interpretation became less meaningful, which is why the nearest neighbor of the predicted vector was chosen as an action index, without sampling.

\Cref{table:ablations} illustrates a significant decline in the variety of actions attempted by the model under this configuration within the bandit environment, showing that the model concentrated only on a fraction of the action set. Though the final regrets for this model were far from random, we point out that this result is because the model skipped exploration phase and went directly to an exploitation of suboptimal actions. That allowed it not to waste time on even more suboptimal ones via exploration. \Cref{appendix:mse_headless_curves} shows the in-context curves of both models, providing more insight into MSE's effect on model quality.

Conversely, on the Darkroom environment, the loss substitution led to a marginal increase in the amount of attempted actions. However, this did not lead to an increased performance. In fact, the performance of this model on the Darkroom environment was similar to a random behavior. 

In summary, although the employed neural network architecture was capable of learning an improvement operator and demonstrating ICL capabilities, the implemented loss function played a crucial role in the success of this learning process. We emphasize the importance of our design choice to use contrastive loss by showing that a more naive approach of directly copying the action embeddings, as incentivized by MSE loss, resulted in underperformance.




\subsection{Orthonormal Action Embeddings}
In this ablation study, we underscored the significance of employing orthonormal vectors for action embeddings by contrasting them with vectors derived from a standard normal distribution. Our preference for the former stemmed from their ability to simplify the approximation of action probabilities because of their property of independence. Unlike embeddings from a standard normal distribution, orthonormal vectors ensure that assigning a probability weight to one vector does not unintentionally influence another (we illustrate it in \Cref{appendix:act_emb_dependence}). This concept echoes the principle of \textit{superposition}, observed when models incorporate more features than available dimensions, leading to feature interference \cite{elhage2022superposition}.

\Cref{table:ablations} illustrates the consequences of using linearly dependent action embeddings, manifesting as diminished performance across Bernoulli bandits and Darkroom scenarios. Notably, these results bear a striking resemblance to "Prompt Ablation", with both indicating a slight performance dip in Darkroom, a more noticeable decline in Bernoulli Bandits, and a general downtrend as the number of bandit arms grows. This parallel underscores a shared objective between these design choices: refining the model's capacity for precise action selection.

\section{Related Work}

Here, we discuss previous research in adapting RL to environments with variable action spaces. For an extended Literature Review, see \Cref{appendix:related_work}.

Recent research by \citet{chandak2020lifelong, ye2023action} has focused on scenarios where the amount of available actions grows during evaluation by introducing new actions to an existing set. However, their model requires fine-tuning when new actions are introduced, while our model does not require parameter updates. Additionally, our research explores a broader spectrum of dynamic action spaces, encompassing the addition, removal, and substitution of actions. Lastly, we work in a setting where the action set remains constant throughout model evaluation.

\citet{kirsch2022introducing} present the concept of SymLA, a methodology that ensures resilience to changes in input and output sizes and permutations. This is achieved by integrating symmetries into a neural network model by representing each neuron with uniformly structured RNNs and data flow with message-passing connections. However, we suggest a simpler approach by extending the well-known transformer architecture. Additionally, we demonstrate the high performance of our model on more complex environments.

Further developments by \citet{jain2020generalization} include a specific module designed to generate action representations informed by observed trajectories after action application. Additionally, \citet{jain2021know} delve into variable action spaces, placing a specific emphasis on modeling the interconnections among actions to improve the quality of action representations. Our work, however, adopts a more implicit approach to inferring action-related information.

In a similar work, \citet{lu2023structured, kirsch2023towards} employ random projections for action encoding, a technique we also utilize. This method facilitates training across multiple domains with actions of varying sizes. Nonetheless, their focus is predominantly on continuous spaces, while our focus is on discrete action spaces.


To the best of our knowledge, Headless-AD is the first study to explore variable discrete action spaces for in-context reinforcement learning.

\section{Conclusion}
In our work, we have introduced a new architecture that extends Algorithm Distillation (AD) for environments with variable action spaces, achieving invariance to their structure and size. Our approach consists of discarding the last linear layer, granting invariance to the action space structure, and making the model infer action semantics from the context, preparing it for the introduction of novel actions. We demonstrated Headless-AD's capability to generalize across new action spaces on a set of environments. We also observed its performance gains over vanilla AD, especially on larger action spaces. We hypothesize that this is due to the augmentation of the dataset by random embeddings, which was shown to improve the generalization abilities of agents \cite{kirsch2023towards}. Headless-AD marks the progress toward versatile foundational models in RL, ones capable of operating across an expanded range of environments. We hope that Headless-AD inspires further development of models that can adapt to any action space beyond discrete ones.

\textbf{Limitations.} Headless-AD shares AD's limitation of fixed sequence lengths, which may limit its effectiveness in environments with long episodes. A unique constraint of Headless-AD is its limit on the number of actions, dictated by the dimensionality of the action embeddings -- only as many actions as there are dimensions can be orthogonally represented. Going over this limit causes embeddings to become linearly dependent, unintentionally distributing probability across multiple actions. While action spaces in RL environments typically do not become excessively large, and increasing the dimension of embeddings could mitigate this issue, it remains a point for consideration.

\textbf{Future Work.} In our study, we demonstrated the ability of our algorithm to generalize to new action spaces, as shown by its performance on elementary tasks. To extend and validate these findings, future research should focus on more complex environments. This will offer a deeper insight into the algorithm's versatility and robustness in diverse and complex settings.

Additionally, to ensure the broader applicability and adaptability of our approach, it is essential to examine its compatibility and performance with various models beyond Algorithm Distillation (AD) \citep{laskin2022context}, such as the Decision Pretrained Transformer (DPT) \citep{lee2023supervised}. This exploration will provide a more comprehensive understanding of the algorithm's strengths and limitations, potentially leading to further improvements and a wider scope of applications in different contexts. 
\section*{Impact Statement}

Strong safeguards are necessary to prevent unauthorized users from manipulating the model, such as adding harmful action embeddings that could lead to negative outcomes.

Another concern is how the model handles out-of-distribution data. If the model encounters a new action that is significantly different from the training actions, it may take a while to understand its effects. Since our model learns by trying out actions, there is a risk it might perform harmful actions before learning they are inappropriate.

Any application of Headless-AD in real-life scenarios should be aware of these potential risks.


\bibliography{icml2024}
\bibliographystyle{icml2024}

\newpage
\appendix
\onecolumn

\section{Background}
\label{appendix:background}
\textbf{Partially Observable Markov Decision Process.}
A Markov Decision Process (MDP) is defined by a tuple $(S, A, P, R)$, where $s_t \in S$ denotes a state, $a_t \in A$ an action, $p(s' | s_t = s, a_t = a)$ the transition probability from state $s_t$ to $s_{t+1}$ after taking action $a_t$, and $R(s, a)$ the reward for action $a$ in state $s$ \citep{sutton2018reinforcement}. An agent $\pi$ observes the state $s_t$, selects action $a_t \sim \pi(\cdot|s_t)$, and receives the subsequent state $s_{t+1} \sim P(\cdot | s_t, a_t)$ and reward $R(s_t, a_t)$. In POMDPs, the agent receives an observation $o_t$ instead of the full state $s$, which contains partial information about the MDP's real state. In the context of our work, $o_t$ may lack goal information, requiring inference from the agent's memory.

\textbf{Multi-Armed Bandits.}
A Bernoulli multi-armed bandit (MAB) environment consists of $N$ arms $a_i \in A$, each associated with a mean $\mu_i$ \citep{sutton2018reinforcement}. Pulling an arm $a_i$ yields a reward $r_i \sim \text{Bernoulli}(\mu_i)$. The agent's objective is to identify the arm with the highest $\mu_i$. Performance is measured using regret, calculated as $\sum_t(\mu_{a^*} - \mu_{\hat{a}_t})$. Unlike MDPs, MABs lack states. 

In a Contextual MAB, each arm $a$ has a feature vector $x_a \in R^d$ \citep{sutton2018reinforcement}. At each step $t$, the agent observes a context state $s_t$. Selecting action $a_i$ results in a reward from a normal distribution with mean $\mu_t = \langle s_t, a_i \rangle$ and standard deviation $\sigma$. Here, unlike in MDPs, the actions influence immediate rewards but not future states.




\textbf{In-Context Learning.} In-Context Learning describes the capability of a model to infer its task from the context it is given. For instance, the GPT-3 model \citep{brown2020language} can be prompted in natural language to perform a variety of functions such as text classification, summarization, and translation, despite not being explicitly trained for these specific tasks. One of the possible prompts, which is used in our paper in some form, is a list of example pairs $(x_i, y_i)$ ending with a query $x_q$ for which the model is expected to generate a corresponding prediction $\hat{y}_q$.

\textbf{Contrastive Learning.} Contrastive learning focuses on creating representations where “similar” examples are close together in the feature space, while “dissimilar” ones are far apart \citep{weinberger2009distance, schroff2015facenet}. This concept may be encapsulated in a triplet loss formula, where the similarity between an anchor and a positive example is maximized, and that between an anchor and a negative example is minimized, expressed as $L = sim(x, x^+) - sim(x,x^-)$. \citet{oord2018representation} has developed a variant of contrastive loss called InfoNCE. This loss was lately widely adopted for representation learning \citep{jaiswal2020survey}.

\section{Related Work}
\label{appendix:related_work}

\subsection{Transformers in Reinforcement Learning}
According to the survey by \citet{li2023survey}, Transformers \citep{vaswani2017attention} are increasingly utilized in reinforcement learning (RL) for various tasks, including representation learning of individual observations and their histories, as well as model learning, as seen in Dreamer by \citet{hafner2019dream}. In our research, we focus on the application of Transformers in sequential decision-making and in developing generalist agents. Same as AD \citep{laskin2022context}, Headless-AD also utilizes transformers as the base model for our approach.

The incorporation of transformers as a sequence modeling tool in RL began with the Decision Transformer (DT) by \citep{hu2022transforming}, which is trained autoregressively on offline datasets of state, action, and return-to-go tuples. Unlike conventional RL, which focuses on return maximization, DT generates appropriate actions during inference by conditioning on specified return-to-go values. The Trajectory Transformer by \citet{janner2021offline} is an alternative approach using beam search to bias trajectory samples based on future cumulative rewards. Building on this, the Multi-Game Decision Transformer (MGDT) by \citet{lee2022multi} improves upon DT with enhanced transfer learning capabilities for new games, eliminating the need for manual return specification. Similarly, the Switch Trajectory Transformer by \citep{lin2022switch} expands on the Trajectory Transformer to facilitate multi-task training. However, when transitioning to novel tasks, these approaches require fine-tuning the model.


\subsection{Offline Meta-RL}
Traditional RL agents, tailored to specific environments, struggle with novel tasks. In contrast, by training across a diverse array of tasks, Meta-RL equips agents with adaptable exploration strategies and an understanding of common environmental patterns \citep{duan2016rl, wang2016learning, rajeswaran2017towards, zhang2018dissection, team2021open, nikulin2023xlandminigrid}. Offline Meta-RL, a subset of this approach, trains agents solely on pre-existing datasets, without direct environmental interaction. A critical challenge here is MDP ambiguity, where task-specific policies misinterpret data due to dataset biases. \citet{dorfman2021offline} propose a data collection method to mitigate this issue, and suggest an approach that treats Meta-RL as a Bayesian RL problem for optimal exploration in new tasks. \citet{li2020focal} introduce FOCAL, a framework separating task identification from control. However, their method relies on a strict mapping assumption that fails in certain scenarios, such as those with sparse rewards. \citet{li2021provably} enhance FOCAL with an attention mechanism and improved metric learning, showing greater robustness in scenarios with sparse rewards and domain shifts. However, most Offline Meta-RL methods rely on explicit task modeling, which can introduce limiting biases. Alternatively, In-Context Learning implicitly infers tasks from environmental interactions, offering a potentially more flexible approach.

\subsection{In-Context Learning in RL}
In-Context Learning (ICL) is an ability of a pretrained model to adapt and perform a new task given a context with examples $\{x_i, f(x_i)\}^n$, where $f$ is a function that gives ground truth targets \cite{brown2020language, wang2023images}. Previous work on ICL \cite{mirchandani2023large} showed that Large Language Models operate as General Pattern Machines and are able to complete, transform and improve token sequences, even when the sequences consist of randomly sampled tokens. This ability supports our design choice to randomly encode the actions of agents. The research on Transformer Circuits \cite{elhage2021mathematical} gives evidence of Transformers' ability to copy tokens, either literally or on a more abstract level, explaining their ICL capabilities. This is particularly useful for our model, which explicitly incentivizes an abstract copying of the action embeddings.

Efforts to blend In-Context Learning (ICL) abilities of transformers with reinforcement learning (RL) are gaining traction, promising to create adaptable RL agents for real-world scenarios with varying conditions. Attempts to transfer ICL features to RL include Prompt-Based Decision Transformers \citep{xu2022prompting}, which leverage task-specific demonstration datasets as prompts, exhibiting strong few-shot learning without weight updates. \citet{laskin2022context} introduced Algorithm Distillation (AD) that trains a policy improvement operator using data from agent-environment interactions of learning RL algorithm, predicting the next action autoregressively. A key strategy here is across-episode training for capturing policy improvement. \citet{lee2023supervised} developed Decision Pretrained Transformer (DPT) using supervised training with unordered environmental interactions as context to predict optimal actions. Under certain conditions, DPT can be proved to implement posterior sampling, resulting in near-optimal exploration. However, DPT's reliance on an optimal policy during training, not always available, is a limitation, similar to AD's constraint on action space structure, restricting their use as fully generalist agents. \citet{lin2023transformers} provided theoretical analysis of AD and DPT, offering guarantees about learning from base algorithms. \citet{wang2024transformers} also provided a theoretical analysis showing that transformers may implement several RL algorithm in-context. \citet{zisman2023emergence} propose a method for distilling the improvement operator from a demonstrator whose actions are initially noisy, with the noise level decreasing progressively throughout the data collection phase. This approach simplifies the data generation compared to AD as it does not require logging the training process of RL models and relaxes DPT's limitation because the demonstrator may be suboptimal. \citet{kirsch2023towards} demonstrated that data augmentation, through random projection of observations and actions, enhances task distribution and generalization on new domains, boosting ICL capabilities. \citet{raparthy2023generalization} study the effect of model size and dataset properties on the success of in-context learning. Other research, acknowledging Transformers' limitations when it comes to long sequences, explores alternative models such as S4 \citep{gu2021efficiently}. \citet{lu2023structured} adapted S4 for RL, showing that it is capable of surpassing LSTM in performance and Transformers in runtime, indicating potential in RL ICL applications.

In contrast to previous works that tackle changing reward distributions or environment dynamics as changing goals, we expand the application of ICLRL to changing action spaces. Our use of random embeddings, motivated by the preparation of the model to unseen actions, may also improve ICL abilities by data augmentation, as suggested by \citet{kirsch2023towards}. 

\subsection{Discarding the Linear Layer}
The Headless LLM approach, introduced by \citet{godey2023headless}, utilizes Contrastive Learning to eliminate the language head from the model architecture. Rather than generating a probability distribution over tokens, it directly predicts token embeddings. This strategy aims to enhance training and inference speeds by discarding a substantial linear layer. While in the RL context, where action spaces are typically small, this might not significantly impact runtime, it does offer a key advantage, as the model is no longer dependent on the number of actions and can therefore handle variable-length action spaces. Unlike Headless LLM, our approach does not rely on contrastive loss for learning action representations. Instead, we use fixed action embeddings, and the model is tasked to output the embedding of the next action. During inference, an action is chosen from a categorical distribution, where the logits are the similarities between the predicted embedding and available actions. Thus, the role of contrastive loss is to enhance the likelihood of selecting the correct action while diminishing the probabilities of other actions. 

Wolpertinger by \citet{dulac2015deep} also employs a similar approach of removing the linear head to improve the train and inference speeds. The authors suggest associating each action with an embedding and performing the training in a continuous action space. A specific action index is chosen as a nearest neighbor of the predicted embedding, and its effect is treated as environment dynamics. 
Most importantly, this nearest neighbor selection process can be optimized using an approximate algorithm. This leads to logarithmic time complexity, which is particularly advantageous in environments with large action spaces. Thus, we believe that Headless-AD is also capable of performing well on large action spaces, benefiting from runtime gains of approximate nearest neighbor lookup. However, Wolpertinger's usage of fixed action embeddings prevents introduction of new actions, highlighting the significance of Headless-AD's usage of randomized action embeddings.
\section{Algorithm Distillation on Permuted Train Sets}
\label{appendix:ad_perm}

\begin{figure}[h]
    \vskip 0.2in
    \centering
    \includegraphics[width=0.35\columnwidth]{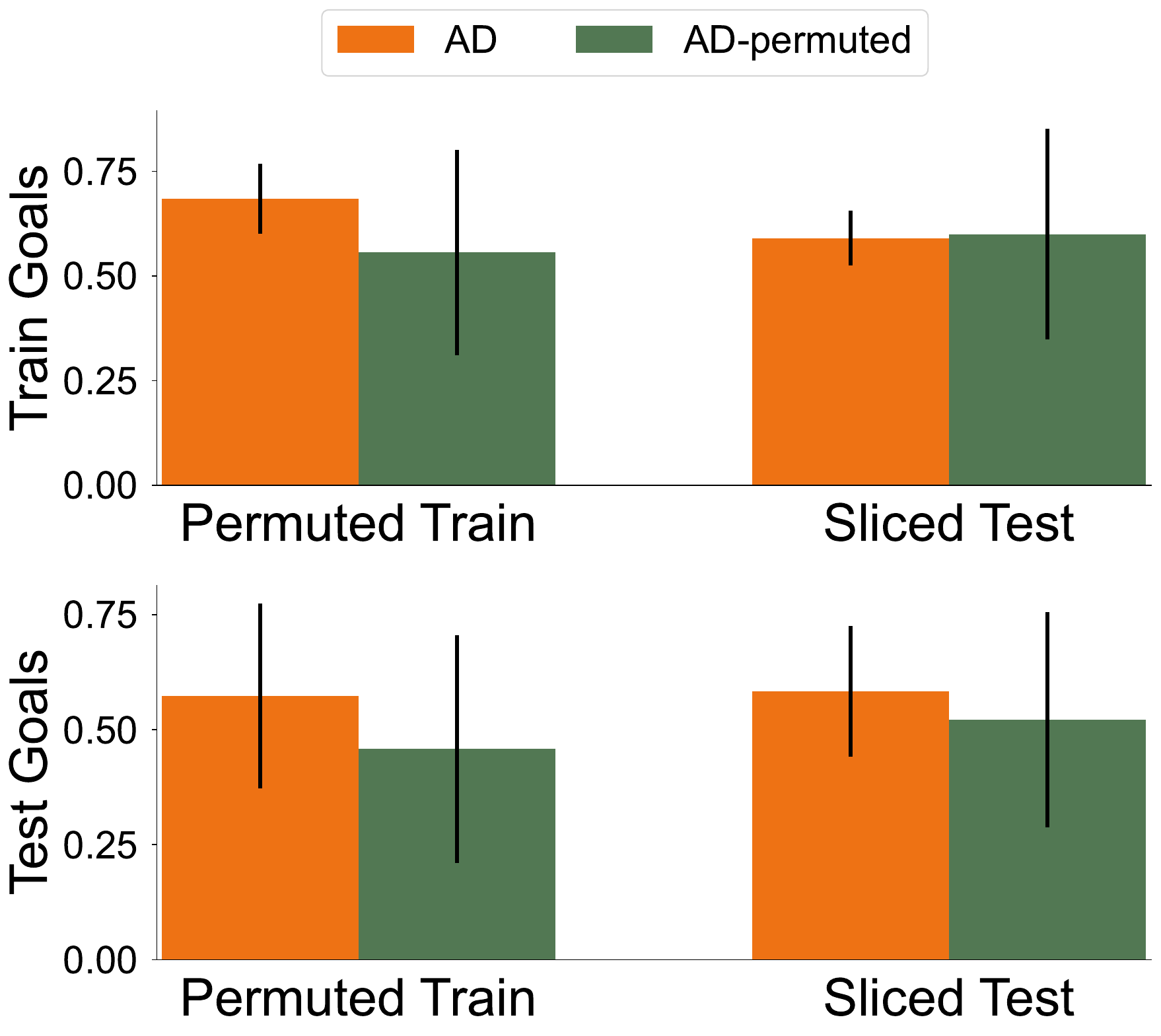}
    \caption{\textbf{AD Performance on Permuted Actions:} This figure shows the models' success rates on Darkroom environment, averaged over $5$ seeds. We evaluated AD's ability to adapt to action sets with varied semantics by training it on a permuted dataset. Except for this, the training and testing are the same as in the vanilla setting and can be found in \Cref{sec:exp_gridworld}. Contrary to expectations, this tailored training did not enhance performance when compared to the AD model trained on standard datasets.}
    \label{fig:gridworld_vanilla_perm}
    \vskip -0.2in
\end{figure}
\section{Across-Environment Generalization}
\label{appendix:cont_to_bern_headless}

We evaluated Headless-AD's ability to generalize skills learned in one domain to another.
We trained Headless-AD on Contextual Bandits and subsequently assessed its performance on Bernoulli Bandits. To align with the Contextual Bandit setting, we introduced a random vector to serve as the state in the Bernoulli Bandit environment, effectively transforming it into a single-context Contextual Bandit scenario with Bernoulli-distributed rewards.

\begin{figure*}[h]
    \vskip 0.2in
    \centering
    \includegraphics[width=0.7\textwidth]{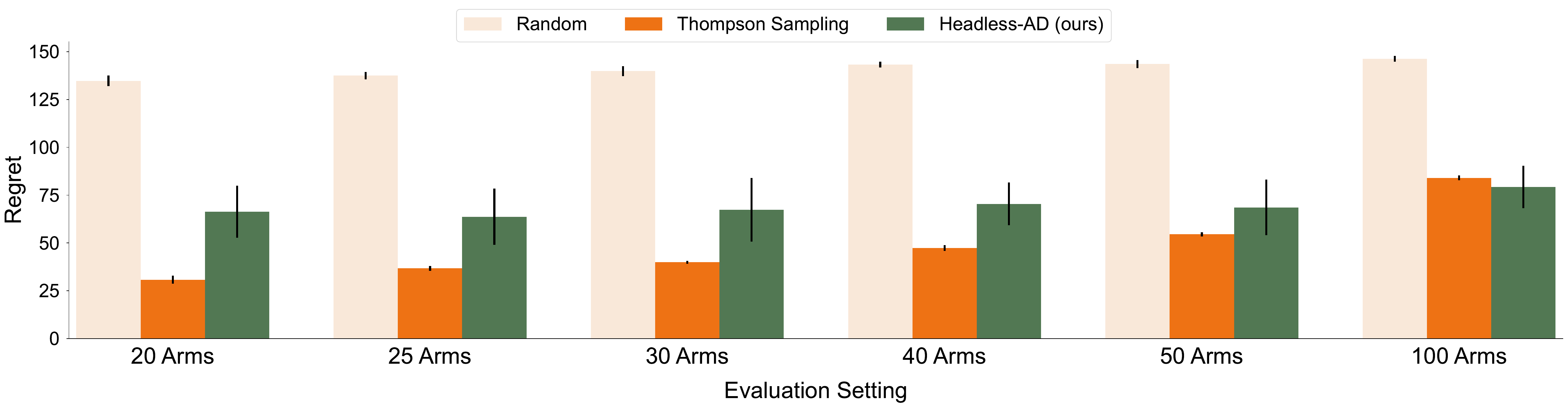}
    \caption{\textbf{Evaluation of Across-Environment Generalization:} This graph illustrates the performance of the Headless-AD model, which was initially trained on a Contextual Bandit setting and then evaluated on a Bernoulli Bandit environment. The experiment aimed to assess the model's generalization capabilities across novel environments. As shown, Headless-AD achieves a decent performance across various configurations with different numbers of arms, demonstrating its potential for across-environment usage.}
    \label{fig:cont_to_bern_headless}
    \vskip -0.2in
\end{figure*}

\Cref{fig:cont_to_bern_headless} presents the results of these cross-domain experiments. Headless-AD not only adapted to the new domain but also maintained a satisfactory performance level, highlighting the model's capacity for cross-environment application.

\section{Visual Darkroom}
\label{appendix:visual_gridworld}
\begin{figure*}[h]
    \vskip 0.2in
    \centering
    \includegraphics[width=0.2\textwidth]{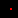}
    \caption{\textbf{Observation in the Visual Darkroom Environment.}}
    \label{fig:visual_gridworld_img}
    \vskip -0.2in
\end{figure*}

We adapted the Darkroom environment to produce high-dimensional visual observations to demonstrate Headless-AD's capability to generalize to new action spaces in more complex settings. In this modified version, the grid is visually represented, with the agent's position indicated in red (see \Cref{fig:visual_gridworld_img}). This change introduces a more complex observation space. For the model to handle these high-dimensional visual observations, we integrated a convolutional network that converts the visual input into an embedding of size 'token\_dim'.

\begin{figure*}[h]
    \vskip 0.2in
    \centering
    \includegraphics[width=0.9\textwidth]{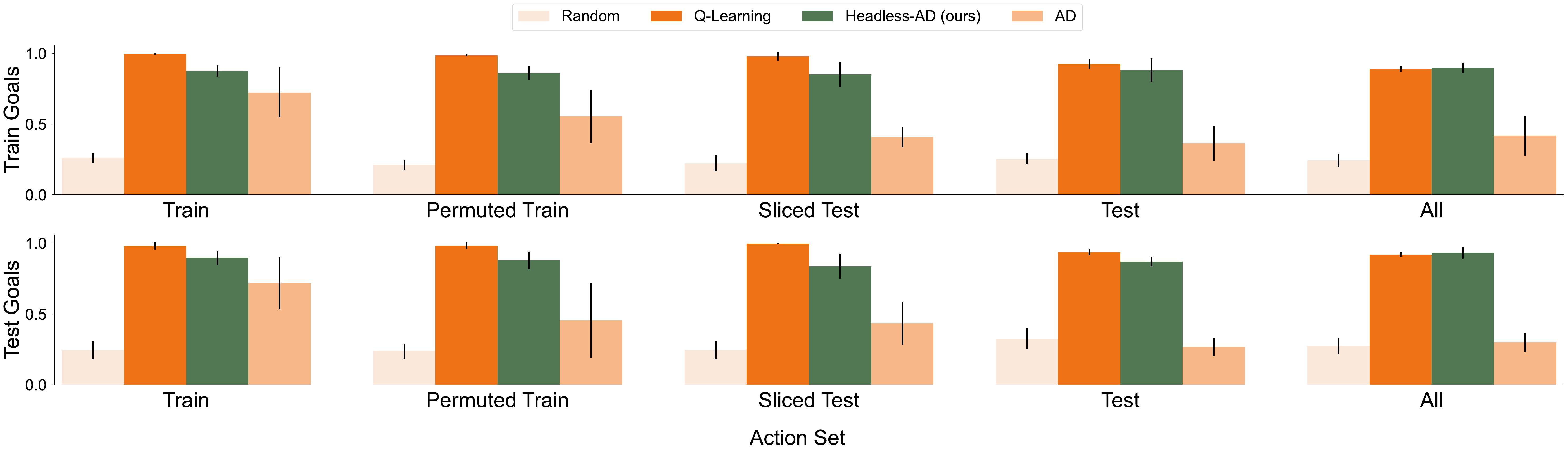}
    \caption{\textbf{Visual Darkroom:} Darkroom environment was modified to emit visual observations. Headless-AD consistently outperforms AD in this more complex observational setting.}
    \label{fig:visual_gridworld_bars}
    \vskip -0.2in
\end{figure*}

Both Headless-AD and AD were configured with the identical hyperparameters previously applied in the Darkroom experiments. The results at \Cref{fig:visual_gridworld_bars} show that Headless-AD maintains its superior performance in this more complex observational setting, whereas AD experiences a notable drop in performance. This discrepancy likely stems from our decision to retain the original hyperparameters without tuning them for the new environment. Based on our experience, AD's performance is particularly sensitive to hyperparameter settings.

\section{Darkroom. Alternative Split}
\label{appendix:gridworld_alternative_split}
A random split of actions in Darkroom environment may lead to a data leakage as multiple action sequences result in  the same endpoint. To address this, we conducted an experiment with distinct, non-overlapping action sets for training and testing. We trained Headless-AD with actions that cause movements of $0$, $1$, and $3$ cells, and tested on movements of $2$ cells, introducing scenarios not encountered during training.

\begin{figure*}[t]
    \vskip 0.2in
    \centering
    \includegraphics[width=0.9\textwidth]{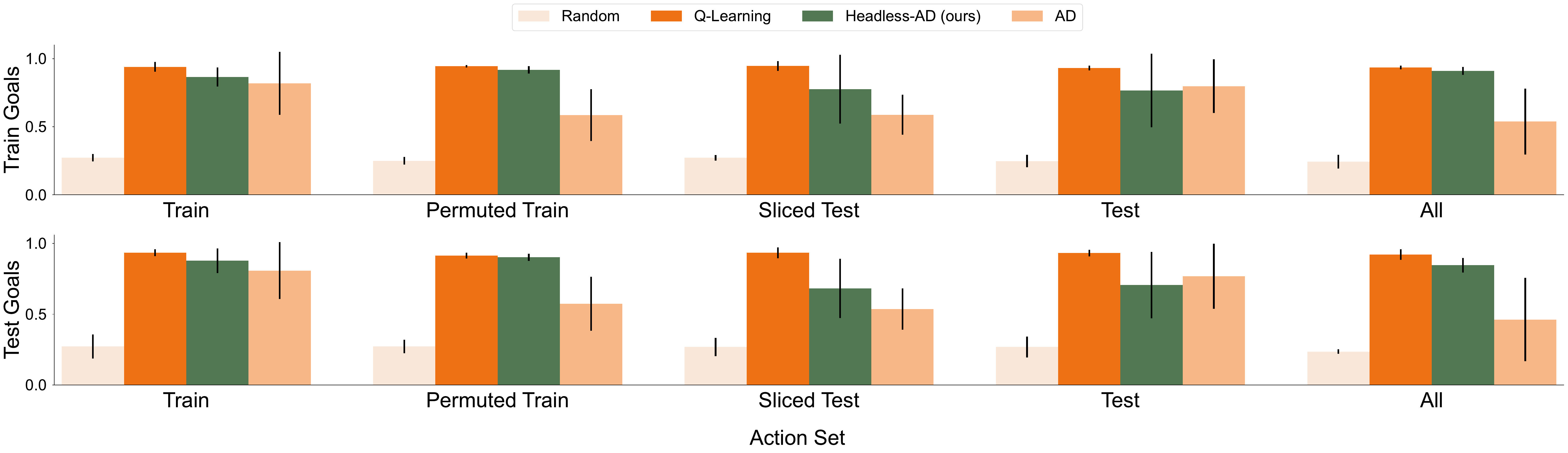}
    \caption{\textbf{Darkroom. Alternative Split:} In this experiment the actions in Darkroom were split into non-overlapping in terms of the distance of the endpoint from the initial position. Train split consisted of actions with length $0$, $1$ and $3$, test split - $2$. Headless-AD, though with a slight drop, maintained its performance on levels seen during the previous experiment with a random split.}
    \label{fig:split_3_gridworld_bars}
    \vskip -0.2in
\end{figure*}

\Cref{fig:split_3_gridworld_bars} shows that although Headless-AD experienced a slight decrease in performance on the test action set, it still remained significantly above the random level and outreaching the AD-level, underscoring its capability to generalize beyond the training action space. This experiment intends to make the capabilities and limitations of Headless-AD clearer.

\vphantom{loh}\\
\vphantom{loh}\\
\section{MSE-Headless-AD In-Context Curves on Bernoulli Bandit}
\label{appendix:mse_headless_curves}


\begin{figure*}[!h]
    \vskip 0.2in
    \centering
    \includegraphics[width=0.9\textwidth]{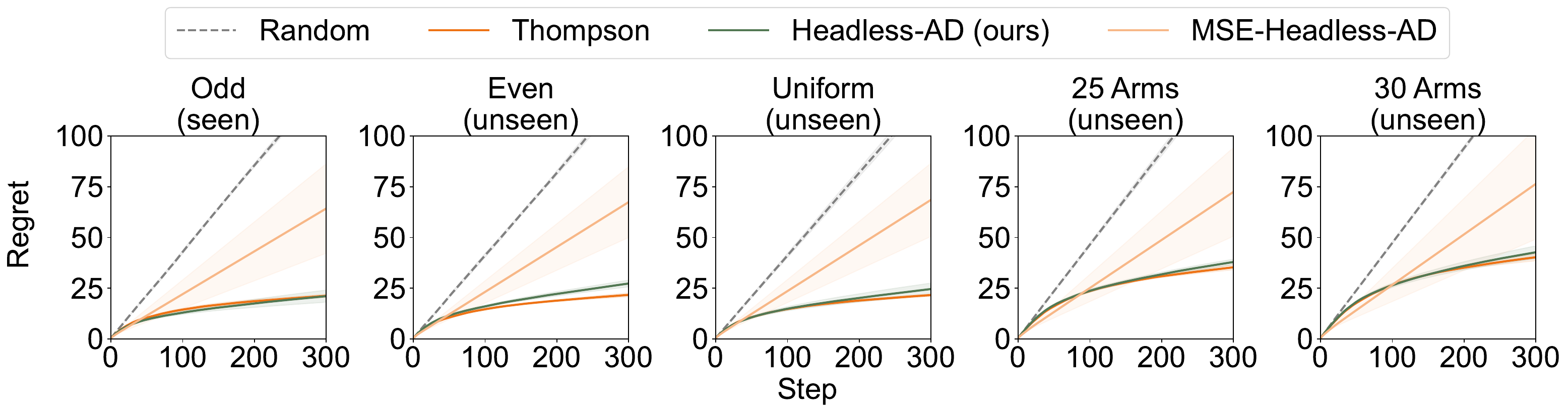}
    \caption{\textbf{MSE-Headless-AD In-Context Curves on Benroulli Bandit:} Though \Cref{table:ablations} shows that a variant of Headless-AD with MSE loss has a far from random regret, this graph shows that this result is due to the exploitation of several actions and the lack of exploration. We make this conclusion by observing that the shape of MSE-Headless-AD's training curve is a straight line, showing the evidence that no learning is being present. All curves are averaged over $5$ seeds.}
    \label{fig:mse_headless_curves}
    \vskip -0.2in
\end{figure*}

\section{Sampling of Orthonormal Vectors}
\label{appendix:orthonormal_vectors}
To sample the orthonormal vectors used as action embeddings, we use the \texttt{torch.nn.init.orthogonal} function from PyTorch \citep{NEURIPS2019_9015} that utilizes a specific algorithm from \citet{saxe2013exact}.

\section{Algorithms' Training Times}

\begin{table}[h]
    \caption{\textbf{Algorithm Training Times:} Here we list the training times (in hours) of Headless-AD and AD on each environment. All experiments were performed on A100 GPUs. Note that times here include both training and evaluation steps. Though Headless-AD requires more time for completion, we point out that it is evaluated on more environment sets compared to AD.}
    \label{tab:algo_training_times}
    \centering
    \begin{tabular}{c|c|c}
         & Headless-AD & AD \\
        \hline 
        Bernoulli Bandit & $5.5$ & $1.5$ \\
        Contextual Bandit & $10$ & $7$ \\
        Darkroom & $27$ & $19$ \\
    \end{tabular}
\end{table}
\section{Model Hyperparameters}
\label{appendix:hypers}
In this section, we detail the hyperparameters for our models, each tuned for specific environments and design configurations. The tuning process utilized Bayesian optimization via the wandb sweep tool \citep{wandb}. The optimization objective was chosen to maximize both the performance score achieved by the model and the efficiency in the number of actions utilized. The objective function was structured as follows:

\[
n_a/N + \text{final\_normalized\_return},
\]

where $n_a$ represents the total number of actions attempted by the model during evaluation, and $N$ signifies the amount of possible actions within an environment. Moreover, we took the return from the final episode in evaluation and normalized it to the $[0,1]$ range, with the lower bound $0$ corresponding to the performance of a random agent, and the upper bound $1$ denoting the efficiency of our data generation algorithm. Darkroom environment already has the returns in the range $[0,1]$, so we did not perform normalization for this environment.

\begin{table}[h]
\centering
\caption{\textbf{Headless-AD's Environment-Specific Hyperparameters:} For certain instances, hyperparameters underwent optimization within the specified ranges in the \textit{Sweep Values} column, utilizing the Bayesian search method facilitated by the wandb sweep tool \citep{wandb}. This process was employed to identify the optimal set of hyperparameters for enhanced performance and fair comparisons.} 
\label{table:hypers_headless}
\resizebox{0.9\textwidth}{!}{%

\begin{filecontents*}{hypers/headless_ad.csv}
Bandit,Gridworld,Contextual Bandit
4,4,8
64,64,8
512,512,1024
300,100,300
1.34,7.8,3.0
3.1e-04,1.7e-04,5.0e-05
4.4e-03,1.7e-02,5.0e-02
0.68,0.72,0.5
0.22,0.7,0.3
0.12,0.19,0.25
300,1000,300
sample,sample,mode
\end{filecontents*}
\pgfplotstabletypeset[
    font=\scriptsize,
    my multistyler/.style 2 args={
        @my multistyler/.style={columns/##1/.append style={#2}},
        @my multistyler/.list={#1}
    },
    create on use/row_names/.style={
        create col/set list={
            Number of Layers,
            Number of Heads,
            Model Dim.,
            Sequence Length,
            $\tau$,
            Learning Rate,
            Weight Decay,
            $\beta_1$,
            Attention Dropout Rate,
            Dropout Rate,
            In-Context Episodes,
            Action Selection
            },
    },
    create on use/sweep_values/.style={
        create col/set list={
            \text{[1,2,4,6,8]},
            \text{[2,4,8,16,32,64},
            \text{[128, 256, 512, 1024, 2048]},
            ,
            \text{[1e-5, 10]},
            \text{[1e-5, 1e-2]},
            \text{[1e-6, 5]},
            \text{[0, 1]},
            \text{[0, 1]},
            \text{[0, 1]},
            ,
            \text{[sample, mode]}, 
            },
    },
    columns/row_names/.style={string type},
    every head row/.style={before row=\toprule, after row=\midrule},
    every last row/.style={after row=\bottomrule},
    col sep=comma,
    columns={row_names, Bandit, Gridworld, Contextual Bandit, sweep_values},
    my multistyler={row_names, Bandit, Gridworld, Contextual Bandit, sweep_values}{string type},
    columns/row_names/.append style={column name=Hyperparameter, column type={l}},
    columns/Bandit/.append style={column name=Bernoulli Bandit},
    columns/Gridworld/.append style={column name=Darkroom},
    columns/Contextual Bandit/.append style={column name=Contextual Bandit},
    columns/sweep_values/.append style={column name=Sweep Values}
]{hypers/headless_ad.csv}
}
\vspace{-10pt}
\end{table}

\begin{table}[h]
\centering
\caption{\textbf{AD's Environment-Specific Hyperparameters:} For certain instances, hyperparameters underwent optimization within the specified ranges in the \textit{Sweep Values} column, utilizing the Bayesian search method facilitated by the wandb sweep tool \citep{wandb}. This process was employed to identify the optimal set of hyperparameters for enhanced performance and fair comparisons.} 
\label{table:ad_hypers}
\resizebox{0.9\textwidth}{!}{%

\begin{filecontents*}{hypers/ad.csv}
Bandit,Gridworld,Contextual Bandit
4,2,8
64,8,8
512,512,1024
300,100,300
5.0e-03,5.0e-02,0.2
1.1e-05,4.0e-04,1.3e-05
1.5e-04,8.0e-06,0.25
0.82,0.95,0.58
0.1,3.0e-02,0.28
0.1,1.0e-02,0.45
300,1000,300
sample,mode,mode
\end{filecontents*}
\pgfplotstabletypeset[
    font=\scriptsize,
    my multistyler/.style 2 args={
        @my multistyler/.style={columns/##1/.append style={#2}},
        @my multistyler/.list={#1}
    },
    create on use/row_names/.style={
        create col/set list={
            Number of Layers,
            Number of Heads,
            Model Dim.,
            Sequence Length,
            Label Smoothing,
            Learning Rate,
            Weight Decay,
            $\beta_1$,
            Attention Dropout Rate,
            Dropout Rate,
            In-Context Episodes,
            Action Selection
            },
    },
    create on use/sweep_values/.style={
        create col/set list={
            \text{[1,2,4,6,8]},
            \text{[2,4,8,16,32,64},
            \text{[128, 256, 512, 1024, 2048]},
            ,
            \text{[0, 1]},
            \text{[1e-5, 1e-2]},
            \text{[1e-6, 5]},
            \text{[0, 1]},
            \text{[0, 1]},
            \text{[0, 1]},
            ,
            \text{[sample, mode]}, 
            },
    },
    columns/row_names/.style={string type},
    every head row/.style={before row=\toprule, after row=\midrule},
    every last row/.style={after row=\bottomrule},
    col sep=comma,
    columns={row_names, Bandit, Gridworld, Contextual Bandit, sweep_values},
    my multistyler={row_names, Bandit, Gridworld, Contextual Bandit, sweep_values}{string type},
    columns/row_names/.append style={column name=Hyperparameter, column type={l}},
    columns/Bandit/.append style={column name=Bernoulli Bandit},
    columns/Gridworld/.append style={column name=Darkroom},
    columns/Contextual Bandit/.append style={column name=Contextual Bandit},
    columns/sweep_values/.append style={column name=Sweep Values}
]{hypers/ad.csv}
}
\vspace{-10pt}
\end{table}

\begin{table}[!ht]
\centering
\caption{\textbf{Headless-AD's Environment-Specific Hyperparameters for Prompt Ablation:} For certain instances, hyperparameters underwent optimization within the specified ranges in the \textit{Sweep Values} column, utilizing the Bayesian search method facilitated by the wandb sweep tool \citep{wandb}. This process was employed to identify the optimal set of hyperparameters for enhanced performance and fair comparisons.} 
\label{table:hypers_headless_prompt_ablation}
\resizebox{0.9\textwidth}{!}{%

\begin{filecontents*}{hypers/action_pool_ablation.csv}
Bandit,Gridworld
4,4
64,64
512,512
300,100
0.92,8.5
5.0e-04,3.5e-04
2.6e-05,4.7e-03
0.78,0.99
0.28,0.4
0.16,0.16
300,1000
sample,sample
\end{filecontents*}
\pgfplotstabletypeset[
    font=\scriptsize,
    my multistyler/.style 2 args={
        @my multistyler/.style={columns/##1/.append style={#2}},
        @my multistyler/.list={#1}
    },
    create on use/row_names/.style={
        create col/set list={
            Number of Layers,
            Number of Heads,
            Model Dim.,
            Sequence Length,
            $\tau$,
            Learning Rate,
            Weight Decay,
            $\beta_1$,
            Attention Dropout Rate,
            Dropout Rate,
            In-Context Episodes,
            Action Selection
            },
    },
    create on use/sweep_values/.style={
        create col/set list={
            \text{[1,2,4,6,8]},
            \text{[2,4,8,16,32,64},
            \text{[128, 256, 512, 1024, 2048]},
            ,
            \text{[1e-5, 10]},
            \text{[1e-5, 1e-2]},
            \text{[1e-6, 5]},
            \text{[0, 1]},
            \text{[0, 1]},
            \text{[0, 1]},
            ,
            \text{[sample, mode]}, 
            },
    },
    columns/row_names/.style={string type},
    every head row/.style={before row=\toprule, after row=\midrule},
    every last row/.style={after row=\bottomrule},
    col sep=comma,
    columns={row_names, Bandit, Gridworld, sweep_values},
    my multistyler={row_names, Bandit, Gridworld, sweep_values}{string type},
    columns/row_names/.append style={column name=Hyperparameter, column type={l}},
    columns/Bandit/.append style={column name=Bernoulli Bandit},
    columns/Gridworld/.append style={column name=Darkroom},
    columns/sweep_values/.append style={column name=Sweep Values}
]{hypers/action_pool_ablation.csv}
}
\vspace{-10pt}
\end{table}

\begin{table}[!ht]
\centering
\caption{\textbf{Headless-AD's Environment-Specific Hyperparameters for Loss Ablation:} For certain instances, hyperparameters underwent optimization within the specified ranges in the \textit{Sweep Values} column, utilizing the Bayesian search method facilitated by the wandb sweep tool \citep{wandb}. This process was employed to identify the optimal set of hyperparameters for enhanced performance and fair comparisons.} 
\label{table:hypers_headless_loss_ablation}
\resizebox{0.9\textwidth}{!}{%

\begin{filecontents*}{hypers/mse_ablation.csv}
Bandit,Gridworld
4,4
64,64
512,512
300,100
5.0e-05,1.3e-05
0.17,2.5
0.91,0.75
1.0e-03,8.0e-02
0.14,0.75
300,1000
mode,mode
\end{filecontents*}
\pgfplotstabletypeset[
    font=\scriptsize,
    my multistyler/.style 2 args={
        @my multistyler/.style={columns/##1/.append style={#2}},
        @my multistyler/.list={#1}
    },
    create on use/row_names/.style={
        create col/set list={
            Number of Layers,
            Number of Heads,
            Model Dim.,
            Sequence Length,
            Learning Rate,
            Weight Decay,
            $\beta_1$,
            Attention Dropout Rate,
            Dropout Rate,
            In-Context Episodes,
            Action Selection
            },
    },
    create on use/sweep_values/.style={
        create col/set list={
            \text{[1,2,4,6,8]},
            \text{[2,4,8,16,32,64},
            \text{[128, 256, 512, 1024, 2048]},
            ,
            \text{[1e-5, 1e-2]},
            \text{[1e-6, 5]},
            \text{[0, 1]},
            \text{[0, 1]},
            \text{[0, 1]},
            ,
            , 
            },
    },
    columns/row_names/.style={string type},
    every head row/.style={before row=\toprule, after row=\midrule},
    every last row/.style={after row=\bottomrule},
    col sep=comma,
    columns={row_names, Bandit, Gridworld, sweep_values},
    my multistyler={row_names, Bandit, Gridworld, sweep_values}{string type},
    columns/row_names/.append style={column name=Hyperparameter, column type={l}},
    columns/Bandit/.append style={column name=Bernoulli Bandit},
    columns/Gridworld/.append style={column name=Darkroom},
    columns/sweep_values/.append style={column name=Sweep Values}
]{hypers/mse_ablation.csv}
}
\vspace{-10pt}
\end{table}

\begin{table}[!ht]
\centering
\caption{\textbf{Headless-AD's Environment-Specific Hyperparameters for Action Embeddings Ablation:} For certain instances, hyperparameters underwent optimization within the specified ranges in the \textit{Sweep Values} column, utilizing the Bayesian search method facilitated by the wandb sweep tool \citep{wandb}. This process was employed to identify the optimal set of hyperparameters for enhanced performance and fair comparisons.} 
\label{table:hypers_headless_embed_ablation}
\resizebox{0.9\textwidth}{!}{%

\begin{filecontents*}{hypers/normal_rand_emb.csv}
Bandit,Gridworld
4,4
64,64
512,512
300,100
1.1e-03,6.6e-04
1.2e-03,9.1e-04
0.94,0.6
0.62,0.53
0.17,3.0e-02
300,1000
sample,sample
\end{filecontents*}
\pgfplotstabletypeset[
    font=\scriptsize,
    my multistyler/.style 2 args={
        @my multistyler/.style={columns/##1/.append style={#2}},
        @my multistyler/.list={#1}
    },
    create on use/row_names/.style={
        create col/set list={
            Number of Layers,
            Number of Heads,
            Model Dim.,
            Sequence Length,
            Learning Rate,
            Weight Decay,
            $\beta_1$,
            Attention Dropout Rate,
            Dropout Rate,
            In-Context Episodes,
            Action Selection
            },
    },
    create on use/sweep_values/.style={
        create col/set list={
            \text{[1,2,4,6,8]},
            \text{[2,4,8,16,32,64},
            \text{[128, 256, 512, 1024, 2048]},
            ,
            \text{[1e-5, 1e-2]},
            \text{[1e-6, 5]},
            \text{[0, 1]},
            \text{[0, 1]},
            \text{[0, 1]},
            ,
            , 
            },
    },
    columns/row_names/.style={string type},
    every head row/.style={before row=\toprule, after row=\midrule},
    every last row/.style={after row=\bottomrule},
    col sep=comma,
    columns={row_names, Bandit, Gridworld, sweep_values},
    my multistyler={row_names, Bandit, Gridworld, sweep_values}{string type},
    columns/row_names/.append style={column name=Hyperparameter, column type={l}},
    columns/Bandit/.append style={column name=Bernoulli Bandit},
    columns/Gridworld/.append style={column name=Darkroom},
    columns/sweep_values/.append style={column name=Sweep Values}
]{hypers/normal_rand_emb.csv}
}
\vspace{-10pt}
\end{table}



\clearpage
\section{Linear Dependence of Different Types of Action Embeddings}
\label{appendix:act_emb_dependence}

\begin{figure*}[h]
    \vskip 0.2in
    \centering
    \begin{subfigure}{0.58\textwidth}
        \centering
        \includegraphics[width=\linewidth]{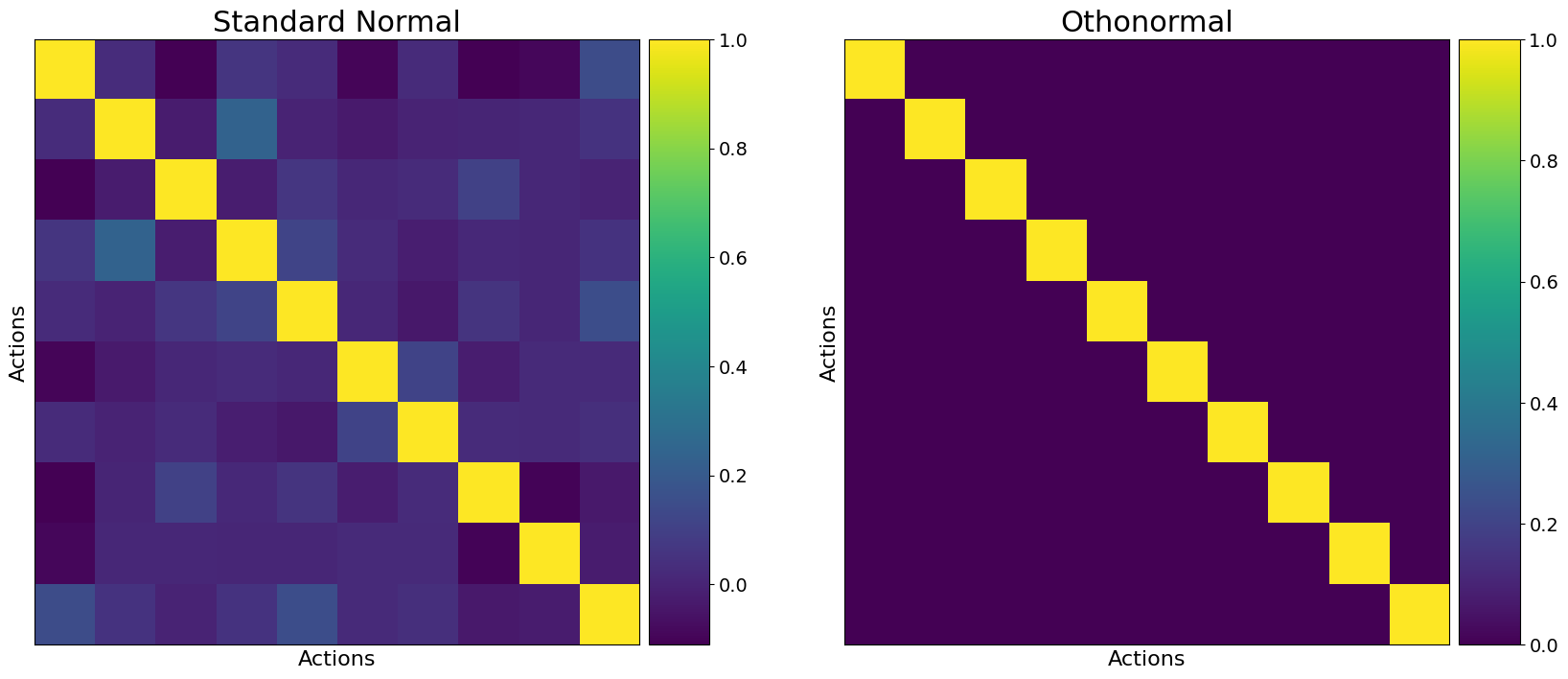}
        \caption{}
        \label{fig:act_sims_128}
    \end{subfigure}
    \begin{subfigure}{0.58\textwidth}
        \centering
        \includegraphics[width=\linewidth]{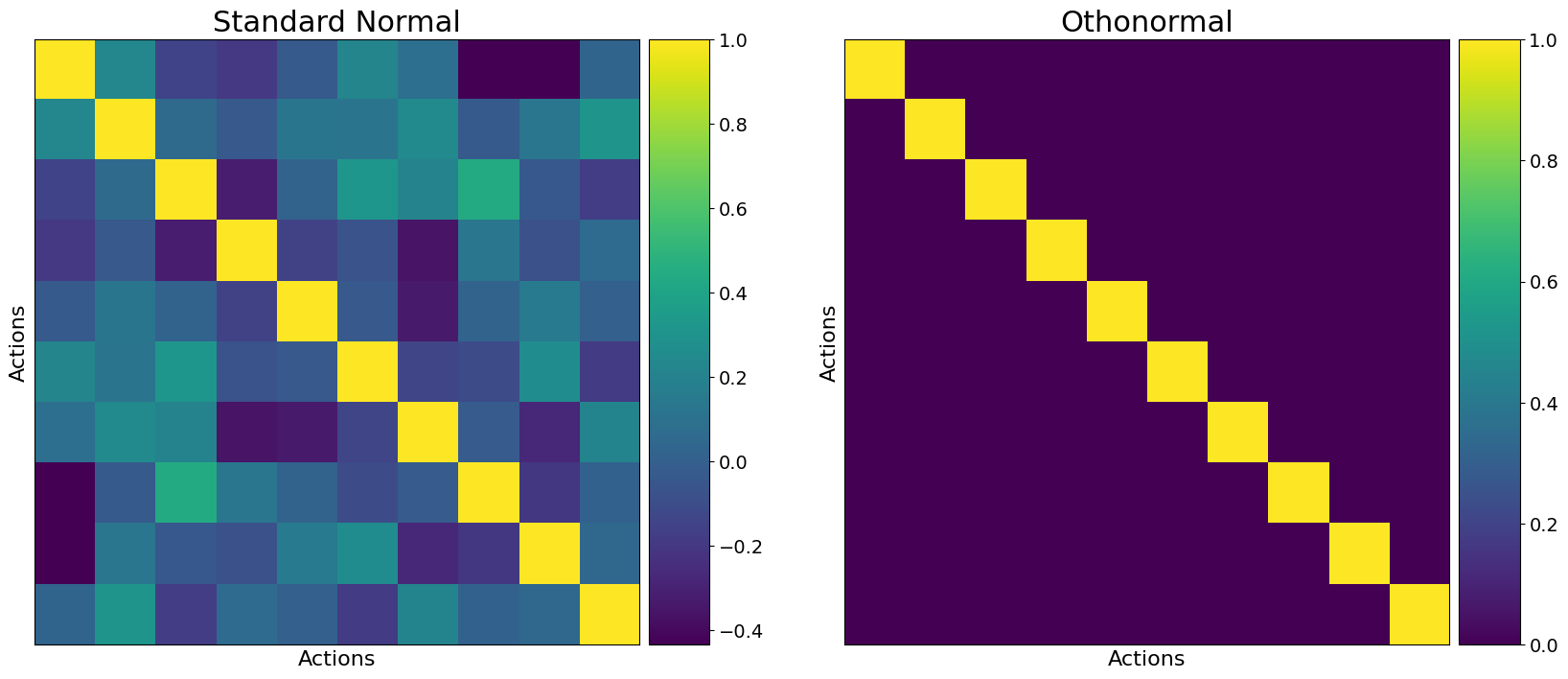}
        \caption{}
        \label{fig:act_sims_16}
    \end{subfigure}
    \begin{subfigure}{0.58\textwidth}
        \centering
        \includegraphics[width=\linewidth]{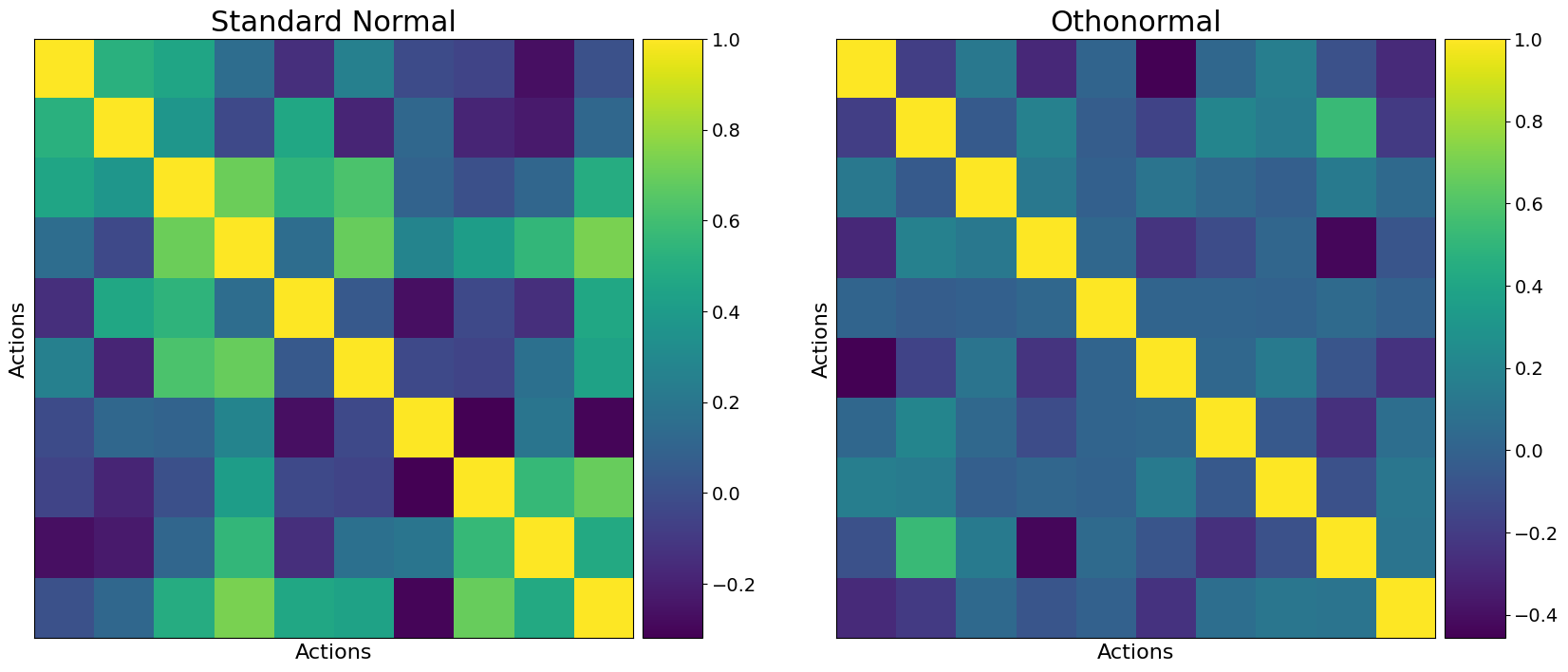}
        \caption{}
        \label{fig:act_sims_more}
    \end{subfigure}
    \caption{These plots illustrate the pairwise cosine similarities between action embeddings for orthonormal versus standard normal distributions. (a) In a $128$-dimensional space with $10$ actions, orthonormal embeddings exhibit perfect decorrelation, whereas standard normal embeddings display non-zero similarities among them. (b) With $10$ actions in a $16$-dimensional space, the similarity between embeddings increases as the dimensionality of the space decreases, indicating denser correlations. (c) For $10$ actions in an $8$-dimensional space, despite the action count exceeding the space's dimensionality, orthonormal vectors (referenced in Appendix: Orthonormal Vectors) maintain lower similarities compared to those from a standard normal distribution, emphasizing the robustness of orthonormal embeddings in constrained dimensions.}
    \label{fig:act_sims_types}
    \vskip -0.2in
\end{figure*}

We illustrate how the probability mass can unintentionally be put on unwanted actions through an analysis of cosine similarities between action embeddings, comparing orthonormal vectors to those from a standard normal distribution. \Cref{fig:act_sims_types} demonstrates non-zero cosine similarities for the standard normal distribution, indicating that a vector perfectly aligned with one action embedding may erroneously attribute non-zero probabilities to other actions.

\clearpage
\section{Code Sample}
\label{appendix:pseudo_code}

\definecolor{mygray}{gray}{0.95}
\begin{listing}[h!]
\begin{tcolorbox}[left*=1.5mm, size=fbox, colback=mygray, boxrule=0.1pt]
\begin{minted}[
    fontsize=\tiny,
    fontfamily=tt
]{python}
from typing import Tuple

import torch
import torch.nn as nn


class ActionMapper(nn.Module):
    def __init__(
        self,
        action_embed_dim: int,
        num_actions: int,
        device: str,
    ):
        super().__init__()
        self.action_embed_dim = action_embed_dim
        self.num_actions = num_actions
        self.device = device

        self.action_map = torch.empty((num_actions, action_embed_dim), device=device)
        self.action_map.requires_grad = False

    @torch.no_grad()
    def regenerate(self) -> None:
        torch.nn.init.orthogonal_(self.action_map, gain=1)

    @torch.no_grad()
    def __call__(self, actions: torch.Tensor) -> torch.Tensor:
        embeds = self.action_map[actions]

        return embeds

    @torch.no_grad()
    def get_action(
        self,
        embeds: torch.Tensor,
    ) -> Tuple[torch.Tensor, torch.Tensor]:
        sims = embeds @ self.action_map.T

        dist = torch.distributions.Categorical(logits=sims)
        actions_sample, actions_argmax = dist.sample(), dist.probs.argmax(-1)

        return actions_sample, actions_argmax

    @torch.no_grad()
    def _get_action_map_as_context(self, batch_size: int):
        actions = torch.tile(self.action_map.unsqueeze(0), (batch_size, 1, 1))

        return actions


act_mapper = ActionMapper(...)

for states, actions, rewards in dataloader:
    act_mapper.regenerate()
    action_embeds = act_mapper(actions)

    actions_list = act_mapper._get_action_map_as_context(batch_size=states.shape[0])

    pred = model(
        states=states,
        act_emb=action_embeds,
        rewards=rewards,
        actions_list=actions_list,
    )

    act_sample, act_mode = act_mapper.get_action(pred)
\end{minted}
\end{tcolorbox}
\label{fig:code_sample}
\caption{Code that demonstrates the Headless-AD training procedure. Note that this snippet is intended for illustration purposes only. The complete code can be found in Headless-AD's repository.}
\end{listing}

\end{document}